\documentclass[10pt]{njulink}
\usepackage{latexsym,makecell,xspace,listings,longtable,needspace,flafter,wrapfig,ragged2e}
\newenvironment{wraptext}{\begingroup\RaggedRight}{\par\endgroup}
\njulinksetup{
  type={Preprint},
  short-title={CEGame-Bench: Long-Horizon Game Creation Agents}
}
\lstdefinestyle{prompt}{
  basicstyle=\normalfont\footnotesize,
  columns=fullflexible,
  keepspaces=true,
  breaklines=true,
  breakatwhitespace=true,
  breakindent=0pt,
  rulecolor=\color{black!25},
framerule=0.4pt,
framesep=5pt,
captionpos=t,
  frame=leftline,
  xleftmargin=4pt,
  xrightmargin=4pt,
  aboveskip=6pt,
  belowskip=4pt,
  showstringspaces=false,
  showlines=true,
  literate={’}{{'}}1 {‘}{{`}}1 {“}{{``}}1 {”}{{''}}1
           {–}{{--}}1 {—}{{---}}1 {→}{{$\rightarrow$}}1
           {…}{{\ldots}}1 {•}{{\textbullet}}1 {×}{{$\times$}}1
           {≥}{{$\geq$}}1 {≤}{{$\leq$}}1
}
\newcommand{\shot}[2]{%
  \IfFileExists{#2}%
    {\includegraphics[width=#1,height=1.25in,keepaspectratio]{#2}}%
    {\fbox{\parbox[c][1.25in][c]{\dimexpr#1-2\fboxrule-2\fboxsep\relax}{\mbox{}}}}}
\definecolor{ogblue}{RGB}{31,119,180}
\definecolor{ogorange}{RGB}{230,120,30}
\definecolor{ogteal}{RGB}{26,150,130}
\definecolor{ogA}{RGB}{117,84,160}
\definecolor{ogS}{RGB}{31,119,180}
\definecolor{ogT}{RGB}{44,160,140}
\definecolor{ogD}{RGB}{230,120,30}
\definecolor{ogH}{RGB}{200,55,60}
\newcommand{\claudemodel}{Claude Opus 5}
\newcommand{\kimimodel}{Kimi K3}

\title{From Code to Playable Systems:\\
CEGame-Bench for Long-Horizon\\
Game Creation Agents}
\author[1,*]{Zixuan~Dong}
\author[2,*]{Yanghai~Wang}
\author[2]{Hanyan~Bian}
\author[2]{Jiaxi~Tao}
\author[2]{Junze~Li}
\author[3]{Jiadong~Wang}
\author[4]{Xinyu~Che}
\author[1]{Lin~Liu}
\author[5]{Baoyun~Peng}
\author[1]{Xiaodong~Wang}
\author[2]{Jiaheng~Liu}
\affiliation[1]{National~University~of~Defense~Technology}
\affiliation[2]{Nanjing~University}
\affiliation[3]{The~University~of~Tokyo}
\affiliation[4]{Xi'an~Jiaotong~University}
\affiliation[5]{Academy~of~Military~Sciences}
\contribution[*]{Equal contribution.}
\metadata[Dataset]{https://github.com/ZexDong/CEGame}
\metadata[Contact]{\email{dongzixuan18@nudt.edu.cn}\quad \email{liujiaheng@nju.edu.cn}}

\newcommand{\bench}{\textsc{CEGame-Bench}\xspace}
\newcommand{\benchfull}{\textbf{C}ross-\textbf{E}ngine \textbf{Game} Bench}

\newcommand{\iface}{\ensuremath{\Phi}}

\newcommand{\xmark}{\tikz[baseline=-0.24ex,x=1.35ex,y=1.35ex]{%
  \draw[black,line width=0.75pt,line cap=round]
    (-0.42,-0.34) .. controls (-0.20,-0.17) and (0.06,0.16) .. (0.39,0.44);
  \draw[black,line width=0.75pt,line cap=round]
    (-0.35,0.44) .. controls (-0.12,0.23) and (0.19,-0.08) .. (0.43,-0.37);}}
\newcommand{\cmark}{\textcolor{blue!80!black}{\ensuremath{\checkmark}}}

\newcolumntype{P}[1]{>{\raggedright\arraybackslash}p{#1}}

\renewcommand{\arraystretch}{1.16}

\abstract{%
A game ships when a player can launch it, finish it, and return to a saved
world, none of which a successful build guarantees, yet game-creation
agents are mostly tested on one engine or on one-prompt browser games.
We present \bench{}, the first long-horizon benchmark for complete-game
creation across five production engines: Defold, Godot, Roblox Studio,
Unity, and Unreal Engine. Agents build each game from an empty project
and a design document in runs averaging up to 295 steps and 65M tokens;
the document doubles as a contract, and a hidden evaluator accepts a game
only by real-input play. Under four scaffolds, agents earn most
integration points but usually under half the gameplay points, and their
games reach the ending in only 16\% of runs, while references built by
engineers with coding agents score near the maximum, which places this
\emph{artifact-to-play gap} in the agents. No two engines rank configurations alike (Kendall's $\tau$ \mbox{0.14--0.62}),
and the same game usually scores higher on Unity and Unreal Engine, which
check more before play. We release Game Completion Boost, an agent skill
for requirement-driven playtesting and repair that raises the
execution-based score of three models by \mbox{14.1--24.3} points.
}

\begin{document}

\maketitle
\suppressfloats[t]

\section{Introduction}
\label{sec:intro}
\begin{figure}[!t]
\captionsetup{font=small}
    \centering
    \includegraphics[width=\linewidth]{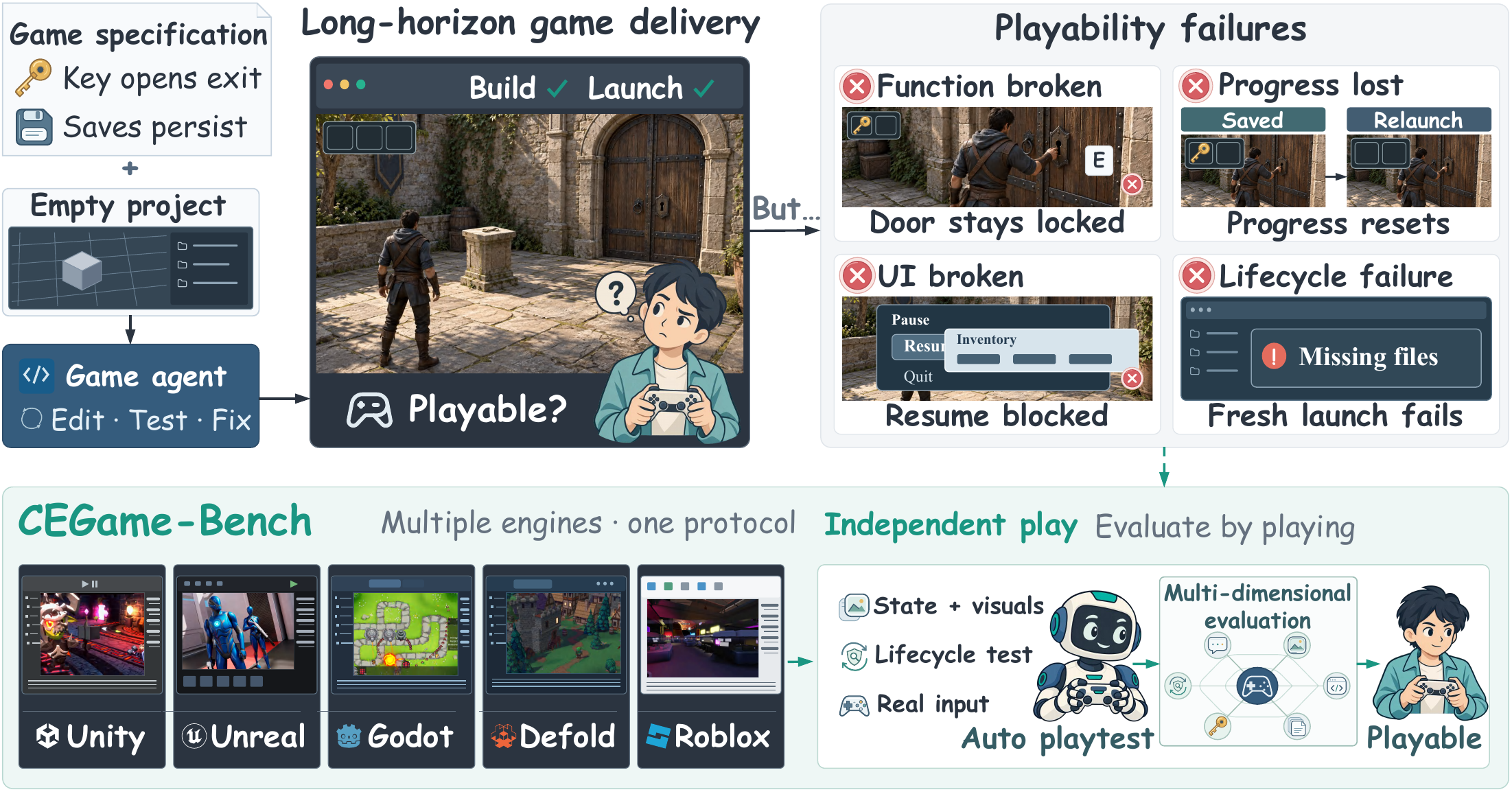}
    \caption{The artifact-to-play gap. \bench{} plays each delivered game through complete player journeys on five production engines.}
    \label{fig:motivation}
\end{figure}

In a game studio, a build ships only after QA plays it to the ending,
quits, and finds the save intact. Coding agents are often shown
on one-prompt browser games~\citep{jiang2026opengame}, which rarely test
what makes a game complete: interdependent parts, where a combat fix
must not break restart. We call the distance between a build and a
playable game the \emph{artifact-to-play gap}
(Figure~\ref{fig:motivation}).

Beyond entertainment, Unity and Unreal Engine, the most demanding engines
to drive, power industrial 3D applications~\citep{la2026gameenginebench}
and embodied-AI simulators~\citep{zhong2025unrealzoo} that coding agents
have begun to build~\citep{kang2026simworldstudio}; making games on them
is a step toward building such worlds from specifications. Because studios choose an engine per project, a benchmark must span
engines to distinguish game-making skill from fluency in one toolchain.

Existing benchmarks each cover one slice:
scoped edits to Godot and Unreal
projects~\citep{chi2026gamedevbench,la2026gameenginebench}, browser
games~\citep{jiang2026opengame,zhang2026webgamebench,zhang2026gameasg},
Godot-to-Unity ports~\citep{chen2026swegame}, or
whole games on a single
engine~\citep{luo2026gamecraftbench,yin2026autoue}. None asks an agent to turn an empty
project into a complete game on several production engines, Unity and
Unreal Engine included, or judges all engines by the same rules
(Table~\ref{tab:related-main}).

We introduce \benchfull{} (\bench{}), which holds every engine to one
execution-based standard. Its 50 tasks are studio handoffs, ten on each
of five production engines, from script-centric Defold, Godot, and
Roblox Studio to heavyweight Unity and Unreal Engine. In each task, the
agent builds a complete game from an empty project and a design
document in a fully logged run. The document is a natural-language
contract of checkable rules and a required state interface, so one
hidden evaluator accepts any implementation on any engine only by
playing it with real input.

The gap is large. Autonomous agents earn most integration points but
usually under half the gameplay points, and their games reach the ending
in only 16\% of runs, while references built by engineers with coding
agents score near the maximum. The engine is a first-class variable: no
two engines rank configurations alike (Kendall's $\tau$ 0.14--0.62), and
the same game scores higher on the heavyweight engine in five of six
engine pairs, by up to 21.5 points, in line with its stricter checks
before play. Nor does effort close the gap: nine in ten runs deliver
before the time limit, and neither steps nor tokens track the score.
These failures motivate \emph{Game Completion Boost} (GCB), a
lightweight agent skill that maps requirements to player checkpoints,
compares state with visible outcomes, and replays affected paths after
repair. On high-difficulty games from each track, GCB raises the
execution-based $F$ of three models under Codex CLI by 14.1--24.3
points and their overall score by 14.7--25.8.

We contribute (i) \bench{}, the first long-horizon benchmark for complete-game creation on five production engines, with controlled engine variation; (ii) a contract-driven evaluation that turns a natural-language design document into executable checks, accepts a game only by play, and logs its construction; and (iii) a cross-engine study of six frontier models under four scaffolds, motivating reusable GCB.

\begin{table}[!t]
\caption{Comparison with prior game-creation work. Replay: the evaluator
plays the artifact with input; Hidden eval.: unseen checks decide the
score; Reopen: persistence is checked in a fresh process; Engine swap:
a task is re-run on another engine.}
\centering
\footnotesize
\setlength{\tabcolsep}{2.2pt}
\renewcommand{\arraystretch}{1.16}
\begin{tabularx}{\linewidth}{@{}
  >{\hsize=1.55\hsize\raggedright\arraybackslash}X
  >{\hsize=1.35\hsize\raggedright\arraybackslash}X
  >{\hsize=1.00\hsize\centering\arraybackslash}X
  >{\hsize=0.95\hsize\centering\arraybackslash}X
  >{\hsize=0.75\hsize\centering\arraybackslash}X
  >{\hsize=0.85\hsize\centering\arraybackslash}X
  >{\hsize=0.65\hsize\centering\arraybackslash}X
  >{\hsize=0.90\hsize\centering\arraybackslash}X@{}}
\toprule
\rowcolor{LinkPanel}
\textbf{Benchmark} & \textbf{Artifact} & \textbf{Engine(s)} & \textbf{Horizon} & \textbf{Replay} & \textbf{Hidden eval.} & \textbf{Reopen} & \textbf{Engine swap} \\
\midrule
GameGPT         & generated game    & one setting  & partial    & partial & \xmark  & \xmark  & \xmark \\
GameXpert-Bench & game lifecycle    & one setting  & multi-turn & \cmark  & \cmark  & partial & \xmark \\
\midrule
GameDevBench    & Godot task        & Godot        & task-level & partial & partial & \xmark  & \xmark \\
GameCraft-Bench & playable game     & Godot        & task-level & \cmark  & partial & \xmark  & \xmark \\
SWE-Game        & game, port        & Godot, Unity & task-level & \cmark  & \cmark  & \xmark  & code port \\
GameEngineBench & C++ runtime patch & Unreal       & task-level & \cmark  & \cmark  & \xmark  & \xmark \\
AutoUE          & 3D game           & Unreal       & task-level & partial & \xmark  & \xmark  & \xmark \\
\midrule
OpenGame        & browser game      & Web          & task-level & browser & partial & \xmark  & \xmark \\
WebGameBench    & browser game      & Web          & task-level & browser & partial & \xmark  & \xmark \\
GameASG-Bench   & browser game      & Web          & task-level & browser & \cmark  & \xmark  & \xmark \\
\midrule
\rowcolor{blue!8}
\textbf{\bench} & \textbf{whole game} & \textbf{5 engines} & \textbf{long-horizon} & \cmark & \cmark & \cmark & \cmark \\
\bottomrule
\end{tabularx}
\normalsize

\label{tab:related-main}
\end{table}

\section{Related Work}
\label{sec:related}

\paragraph{Long-horizon software and computer-use agents.}
Repository benchmarks cover issue
resolution~\citep{jimenez2024swebench,deng2025swebenchpro}, feature
evolution~\citep{zhou2026featurebench,le2025sweevo,xu2026roadmapbench},
and construction from
specifications~\citep{ding2026nl2repo,peng2026repogenesis,wang2026codespec};
coding agents~\citep{yang2024sweagent,wang2024codeact,wang2024openhands}
and computer-use
benchmarks~\citep{xie2024osworld,zhu2026workflowgym,ai2026prosoftarena}
add tool use and long workflows. Most score only the final outcome, even though the scaffold alone shifts
accuracy and cost~\citep{kapoor2026hal}. Repository tasks, moreover,
involve no engine serialization, player input, or runtime transitions;
\bench{} involves all three and logs every construction step alongside the
score.

\paragraph{Game-development benchmarks.}
Game benchmarks each cover one engine, the browser, or part of a game:
multi-agent development~\citep{chen2023gamegpt}, scoped
edits~\citep{chi2026gamedevbench,la2026gameenginebench},
Godot~\citep{luo2026gamecraftbench,sun2026jamer} or
Unreal~\citep{yin2026autoue} projects, browser
games~\citep{jiang2026opengame,zhang2026webgamebench,zhang2026gameasg},
generation and repair~\citep{chen2026gamexpert}, and 3D or cinematic
content~\citep{yang2026unigen,he2026cutscene}. SWE-Game alone crosses
engines, but its Unity port starts from existing Godot
code~\citep{chen2026swegame}. On Godot, JAMER finds that code agents
raise compilation rates without improving runtime
behavior~\citep{sun2026jamer}; \bench{} has agents build complete games
from empty projects on five engines, measures this gap through full
player journeys, and re-issues contracts across engines without
porting code.

\paragraph{Executable verification and playability.}
Prior work checks artifacts against executable specifications or hidden
contracts~\citep{jia2026gamegenverifier,lu2026worldcoder}, tests Unity
scenes~\citep{liu2026mage}, and studies GUI play and bug
discovery~\citep{peng2026playcoder,jiang2026gbqa}. GameASG-Bench
declares an evaluation interface for real-input
checks~\citep{zhang2026gameasg} and GameCraft-Bench verifies games by
replayed gameplay~\citep{luo2026gamecraftbench}, each on a single
runtime, while others still grade play with a multimodal
judge~\citep{jiang2026opengame}. \bench{} carries
a declared state interface across five runtimes, scores all but
presentation by execution, and adds fresh-process reopening and mutant
calibration.

\section{Benchmark Design}
\label{sec:benchmark}

\begin{figure}[!t]
    \centering
    \includegraphics[width=\linewidth]{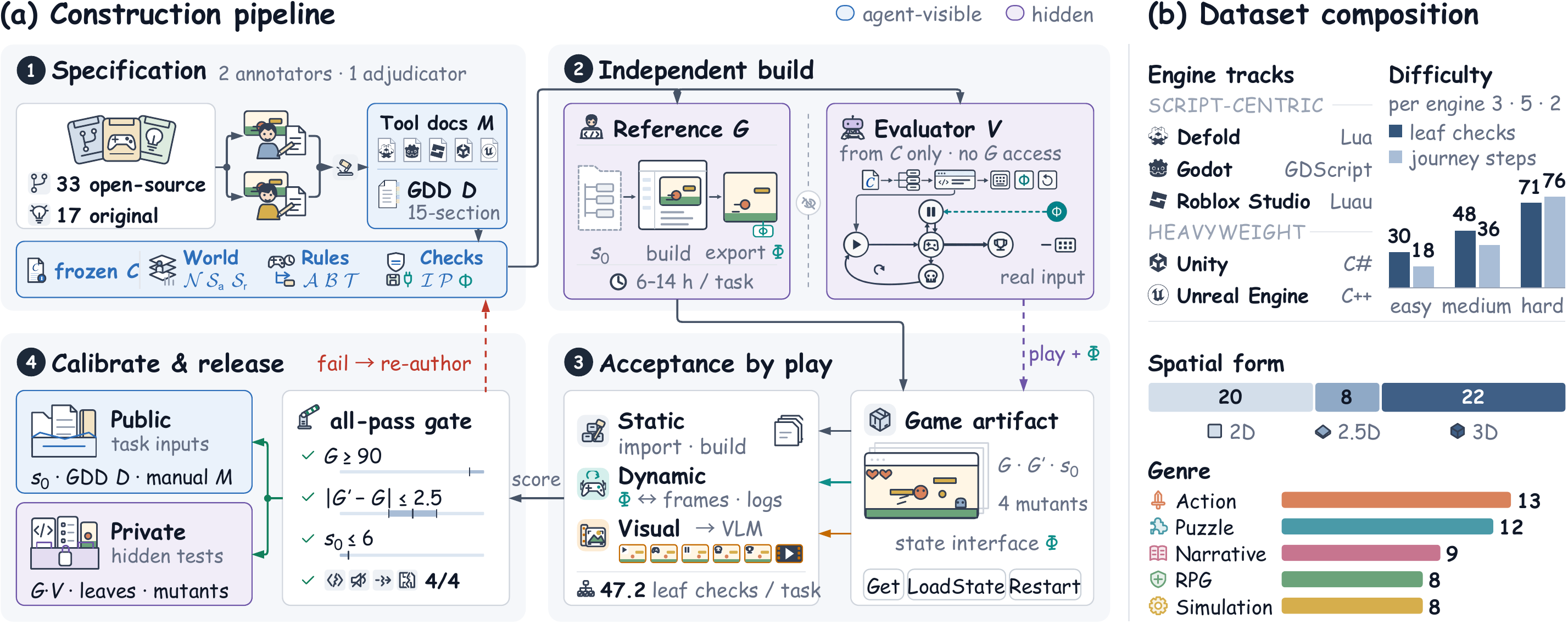}
    \caption{Construction and composition of \bench{}. (a) Tasks are specified as contracts, built, evaluated independently, and calibrated by play. (b) Coverage: engines, difficulty tiers, spatial forms, and genres.}
    \label{fig:pipeline}
\end{figure}

\bench{} turns the studio request \emph{ship this game} into a
long-horizon task anchored by one document: a game design document (GDD)
written as a contract that specifies the game, requires it to expose its
state, and yields the checks a hidden evaluator runs during play. One
contract language and scorecard span five engines, so
the engine is a controlled variable; every task is calibrated on a
reference game and its broken variants, and every run is logged from
first edit to delivery (Figure~\ref{fig:pipeline}).

\subsection{Engine Tracks}
\label{sec:tracks}

The benchmark includes ten tasks on each of five production engines, which
differ in where a game lives (Appendix Table~\ref{tab:engines}). On the
script-centric Defold, Godot, and Roblox Studio, a game is mostly Lua,
GDScript, or Luau code on a scene or instance tree. On the heavyweight
Unity and Unreal Engine, much of it lives in imported assets and
editor-serialized state behind an import and compile cycle that is slow
but validates the game before it runs; Unreal Engine adds binary
\texttt{uasset} files, C++ modules, and build targets. Everything else
is fixed across tracks (Appendix~\ref{app:fairness}): the 3/5/2 mix of
easy, medium, and hard tasks, the starter and GDD structure, an
engine-neutral prompt, the scorecard, and the 3600\,s budget.
Each track ranks configurations on its own games, and a swap experiment
isolates the engine by re-issuing a game's frozen contract on an engine
of the other group.

\subsection{Task Formulation}
\label{sec:task-formulation}

A task comprises $(s_0,D,M,G,V)$. The agent receives a starter project
$s_0$, the GDD $D$, and a tool manual $M$; the reference implementation
$G$ and evaluator $V$ stay hidden. $s_0$ is an engine-native project
that builds, runs, and bundles licensed assets but implements none of
the game, $M$ documents the engine's build, editor-automation, and
capture tools, and $D$ is a fifteen-section GDD from pitch and controls
to persistence and acceptance. From these alone, the agent
works as a studio engineer would: reading documentation, authoring
assets and scripts, compiling, playing, and repairing.

\paragraph{The GDD is a contract.}
The normative core of $D$ is an implementation-independent contract
\[
C=(\underbrace{\mathcal{N},\mathcal{S}_a,\mathcal{S}_r}_{\text{world}},\;
\underbrace{\mathcal{A},\mathcal{B},\mathcal{T}}_{\text{rules}},\;
\underbrace{\mathcal{I},\mathcal{P},\iface}_{\text{guarantees}}),
\]
whose symbols denote entities, authored and runtime state, player
actions, branches, temporal rules, invariants, persistence requirements,
and the state interface. $D$ fixes what the player experiences (levels,
rules, and constants) and the interface through which the game is
observed, and leaves class hierarchy, scene organization, and asset
layout free. Every requirement in $D$ becomes one or more leaf
checks of $V$, so the document that specifies the game also defines its
rubric. In \emph{Driftfolk Chronicle}, for example, ``a dash spends one
charge'' becomes a check on \texttt{player.charges}.

\paragraph{Observability is part of the task.}
Just as a QA build exposes its state to testers, every game must expose
a small state interface $\iface$: a read accessor
\texttt{OmniState.\allowbreak Get(key)} over task-specific keys and two
lifecycle commands, \texttt{OmniFlow.\allowbreak LoadState(id)} and
\texttt{OmniFlow.\allowbreak Restart()}, built on the game's own loading path.
$\iface$ lets the evaluator observe a game without taking its word: it
reads and enters declared states but never acts for the player, every
action arrives as real input, and every reading is cross-checked against
a frame, a log line, or a second key it constrains. Because hidden
probes vary inputs, timing, and rejected actions, a reading stays
correct only if the game computes it from play, so an interface that
reports constants or scripted values fails its own checks
(Appendix~\ref{app:gdd}). With $\iface$, one evaluator
tests any implementation on all five engines without reading its code.

\subsection{Evaluation and Scoring}
\label{sec:scoring}

$V$ accepts a game the way QA does, by playing it. \emph{Static} probes
check import, compilation, the entry scene, serialized references, and
the asset manifest. \emph{Dynamic} probes start from a real entry point
or a state entered through $\iface$, send real input through the
engine's input layer, and assert on $\iface$ snapshots, logs, and frames
after each action, including rejected actions, restart, and
fresh-process reopening. Entering later states directly keeps failures
local: a level that never loads costs only its own leaves. A
\emph{player journey} then plays the whole game with public controls
alone, executing each contract-level step as a closed-loop policy
steered by $\iface$ readings, so differently built games complete the
same path. \emph{Visual} probes score captures at reference states with a
vision--language model that agrees with human raters at Spearman
$\rho=0.81$.

A task decomposes into $47.2$ atomic \emph{leaf checks} on average, in
six dimensions (Appendix~\ref{app:scoring}): integration (D1),
completeness (D2), mechanics (D3), lifecycle (D4), feedback (D5), and
presentation (D6). D1--D5 are scored by execution alone and only D6 by
model judgment (Appendix~\ref{app:stages}). Each leaf $\ell$ has a
weight $w_\ell\in\{1,2,3\}$ (minor, major, critical) and belongs to a
\emph{family} $f$ of leaves expressing the same obligation. With
$\mathcal{L}_f$ the judged leaves of $f$ and $y_\ell$ the credit of
$\ell$, 1 for a pass and 0 for a failure (a D6 leaf takes the judge's
normalized rating),
\[
\pi_f =
\frac{\sum_{\ell\in\mathcal{L}_f} w_\ell y_\ell}
     {\sum_{\ell\in\mathcal{L}_f} w_\ell},
\qquad
S_d = W_d \sum_{f\in\mathcal{F}_d}
      \beta_{d,f}\pi_f,
\]
where $(W_1,\dots,W_6)=(10,30,20,20,10,10)$. Family budgets
$\beta_{d,f}$ sum to one per dimension and stay fixed when a task moves
between engines, so an obligation weighs the same however many leaves it
takes on a track.
The functional score $F=\sum_{d=1}^{5}S_d$ (out of 90) is determined
entirely by execution, and its D2--D5 part (out of 80) gives the
\emph{gameplay points}, earned only by a game that works under play. The
overall score $O=F+\tilde{S}_6$ (out of 100) takes $\tilde{S}_6=S_6$
when the player journey succeeds and $\min(S_6,W_6/2)$ otherwise, so
presentation never outweighs play.

A leaf fails only on evidence: the evaluator observes the violation, or
the submission itself blocks a required state, as a missing $\iface$
blocks every leaf that reads it. Evaluator faults are retried through an
alternate route and, if unresolved, leave the denominator; these
exclusions remove at most 2.9\% of leaf weight on any engine. Beyond the scorecard, each run yields
three binary flags: a completed journey ($J$), state that survives
restart and reopening ($\Lambda$), and a fully \emph{solved} game, in
which both hold and every critical leaf passes. Each run also keeps a
full trace of every tool call, build, and play session with its tokens
and time, so the benchmark reports what each game cost
to build alongside its score.

\subsection{Dataset Construction and Statistics}
\label{sec:construction}
\label{sec:coverage}

Of the 50 tasks, 33 are adapted from permissively licensed open-source
games and 17 are original designs. Each passes four stages
(Appendix~\ref{app:construction}).
(i)~\emph{Specification}: two annotators draft and cross-check the GDD
in player-visible terms, and a third adjudicates and freezes $C$.
(ii)~\emph{Independent build}: engineers working with coding agents
build $G$ from the frozen starter in 6--14 engineer-hours, while $V$ is
written from $C$ alone, without access to $G$. (iii)~\emph{Acceptance by
play}: $V$ plays $G$, an equivalent implementation $G'$ with a
deliberately different structure, the untouched starter $s_0$, and four
mutants that break compilation, feedback, a transition, or persistence,
exactly as it later plays submissions. (iv)~\emph{Calibration gate}: the
task is frozen and released only if, on $O$, $G$ scores at least 90, $G'$
lands within 2.5 points of $G$, $s_0$ scores at most 6, and every mutant
is detected while moving the score by at most 2.0 points outside its
dependency subgraph, so a fault may block what depends on it but nothing
else; otherwise it is re-authored.

The released references anchor the scale. They average $96.4$ on $O$ (minimum $92.5$), the equivalence gap averages $2.0$, starters score $4.1$, all
200 mutants are caught, and repeated evaluations of one artifact differ
by at most $1.8$ points. Since engineers built each reference with
coding agents, these scores mark what agents reach under human
direction, the bar for autonomous agents.

Every document is newly written, and adapted games are renamed and
re-parameterized (Appendix~\ref{app:construction}).
Upon publication we release the starters, design documents, tool
manuals, the evaluator and its adapters, leaf schemas, and run traces,
all carrying a canary string; references, equivalent implementations,
and mutants stay private.

The tasks span 2D, 2.5D, and 3D puzzle, action, RPG, simulation, and
narrative games, with difficulty tiers annotated on four axes
(Appendix~\ref{app:tasks}). A task averages 2.6 scenes, 14.4 entities,
and a 38.6-step player journey; hard tasks average 71 leaf checks and a
76-step journey.

\Needspace{60pt}
\section{Game Completion Boost}
\label{sec:method}

Section~\ref{sec:res-gap} locates failures in completeness, mechanics,
and lifecycle, and Appendix~\ref{app:results} shows unreachable goals,
rules that contradict the scene, and cropped or crowded views.
Game Completion Boost (GCB) targets these failures by organizing
the agent's playtesting and repair
(Figure~\ref{fig:completion-skill}), combining specification-derived
checks~\citep{jia2026gamegenverifier}, file-bound verification
records, and reflective
repair~\citep{shinn2023reflexion}.

\Needspace{160pt}
\begin{wrapfigure}{r}{0.52\textwidth}
    \centering
    \includegraphics[width=\linewidth]{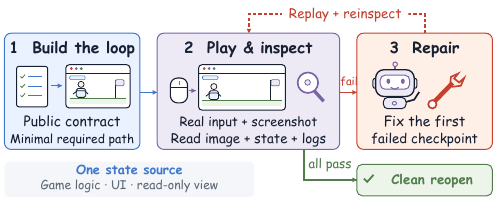}
    \caption{GCB workflow: build the player path, play and inspect,
    then repair, replay, and restore if needed.}
    \label{fig:completion-skill}
\end{wrapfigure}

\begin{wraptext}
\paragraph{(1) Requirement-directed checkpoints.}
The agent maps public requirements, branches, and lifecycle events to
checkpoints with prerequisites, relevant files, and required evidence,
and checks with public controls that required states are reachable.
A helper schedules checkpoints whose prerequisites pass and recommends
repairing failures, gathering missing observations, or replaying stale
records.

\paragraph{(2) Action-linked state--view evidence.}
For visible transitions, the agent compares before/after images with
available state or logs from the same input and session, checking for
disagreements such as a trap firing on visibly safe floor. Static image
checks cover camera framing and UI overlap. Behavior and its
visible presentation both drive repair.

\end{wraptext}

\paragraph{(3) Replay and guarded repair.}
After an edit, GCB invalidates checks tied to changed files and their
dependents; shared or unmapped changes invalidate all checks, following
regression test selection~\citep{gligoric2015ekstazi}. Before repair,
GCB saves the project and protects the passed checkpoints. It retains a
candidate only after the target and all protected checks pass again;
an observed regression triggers rollback. The agent records a diagnosis
to guide its next attempt. Every edit clears final acceptance, requiring
a complete journey and any specified fresh-process reopen.

The released GCB consists of fixed, task-agnostic instructions and
Python helpers. The agent performs and interprets every check from
public inputs within the task budget, and the hidden evaluator alone
scores the submitted game (Appendix~\ref{app:recovery}).

\section{Experiments}
\label{sec:experiments}
\label{sec:results}

Evaluations of game creation commonly take two shortcuts: they treat a
project that builds as a working game, and they measure one engine as
if it stood for all. We test both, measure what each
game costs to build, and evaluate GCB across engine tracks.

\subsection{Setup}
\label{sec:setup}

We evaluate seven configurations pairing six frontier models from four
providers with four scaffolds. Codex CLI runs GPT-6 Astra, GPT-5.6 Sol,
GPT-5.6 Terra, and DeepSeek-V4-Pro, so contrasts within it isolate the
model; DeepSeek-V4-Pro also runs under DeepSeek Harness, isolating the
scaffold; and \claudemodel{} and \kimimodel{} run in their providers'
Claude Code and Kimi Code.

In the main study, each game--configuration pair is run three times from the frozen starter
of its task, with the GDD, tool manual, 3600\,s budget, engine version,
tool catalog, model, and reasoning setting fixed
(Appendix~\ref{app:configs}); the native study comprises 1,050 runs. We
report D1--D6, $F$, and $O$ averaged over the three runs, weighting the
games of a track equally, and compare configurations on $F$, which
execution alone determines. The standard deviation of $F$ across the
three runs of a pair averages 8.4 points, so run-to-run noise moves an
engine cell by about 1.5 points.

\begin{table}[!ht]
\caption{Scores of seven scaffold--model configurations on each engine
track and across all five engines. D1--D6 are integration,
completeness, mechanics, lifecycle, feedback, and presentation, with D6
capped when the player journey fails; $F$ sums D1--D5 (out of 90) and
$O$ adds D6 (out of 100). Bold marks the best $F$ and $O$ per panel;
entries are rounded independently and may differ from their sums by 0.1.}
\centering
\footnotesize
\color{black}
\arrayrulecolor{black}

\colorlet{benchHeader}{LinkPanel}
\definecolor{benchGroup}{gray}{0.95}
\definecolor{benchStripe}{gray}{0.97}

\setlength{\tabcolsep}{1.4pt}
\renewcommand{\arraystretch}{1.03}

\begin{tabular*}{\linewidth}{@{\extracolsep{\fill}}
l @{\hspace{5pt}} l
*{6}{r} @{\hspace{3pt}} rr
@{\hspace{8pt}}
*{6}{r} @{\hspace{3pt}} rr
@{}}
\toprule

\rowcolor{benchHeader}
\multicolumn{2}{c}{\textbf{Configuration}}
& \multicolumn{8}{c@{\hspace{10pt}}}{\textbf{Defold}}
& \multicolumn{8}{c@{}}{\textbf{Godot}} \\
\cmidrule(lr){3-10}\cmidrule(l){11-18}

\rowcolor{white}
Scaffold & Model
& D1 & D2 & D3 & D4 & D5 & D6 & $\mathbf{F}$ & $\mathbf{O}$
& D1 & D2 & D3 & D4 & D5 & D6 & $\mathbf{F}$ & $\mathbf{O}$ \\
\midrule

\rowcolor{white}
\cellcolor{benchGroup}
& GPT-6 Astra
& 10.0 & 20.3 & 11.3 & 14.8 & 6.2 & 9.5 & \textbf{62.7} & \textbf{72.2}
& 9.4 & 15.8 & 8.2 & 12.1 & 4.4 & 5.1
& \textbf{49.8} & \textbf{54.8} \\

\rowcolor{benchStripe}
\cellcolor{benchGroup}
& GPT-5.6 Sol
& 10.0 & 19.0 & 9.8 & 8.7 & 5.0 & 5.8 & 52.5 & 58.3
& 9.6 & 3.8 & 1.9 & 4.9 & 1.3 & 3.1 & 21.5 & 24.6 \\

\rowcolor{white}
\cellcolor{benchGroup}
& GPT-5.6 Terra
& 7.2 & 7.4 & 8.0 & 8.0 & 2.6 & 2.7 & 33.2 & 35.9
& 9.2 & 1.6 & 1.5 & 3.1 & 1.1 & 3.4 & 16.4 & 19.8 \\

\rowcolor{benchStripe}
\cellcolor{benchGroup}\multirow{-4}{*}{Codex CLI}
& DeepSeek-V4-Pro
& 9.5 & 7.2 & 3.7 & 3.7 & 1.8 & 1.8 & 25.8 & 27.6
& 8.1 & 10.2 & 5.4 & 7.2 & 3.3 & 2.7 & 34.2 & 36.9 \\

\rowcolor{white}
DeepSeek Harness & DeepSeek-V4-Pro
& 10.0 & 13.2 & 8.0 & 7.8 & 2.6 & 3.7 & 41.6 & 45.3
& 9.8 & 9.6 & 4.6 & 8.1 & 3.3 & 3.6 & 35.4 & 39.0 \\

\rowcolor{benchStripe}
\cellcolor{benchGroup}Claude Code
& \claudemodel
& 9.5 & 15.7 & 8.6 & 8.3 & 6.7 & 5.9 & 48.8 & 54.7
& 8.0 & 1.5 & 2.0 & 7.0 & 2.9 & 0.5 & 21.4 & 21.9 \\

\rowcolor{white}
Kimi Code & \kimimodel
& 5.8 & 0.0 & 0.0 & 0.0 & 0.0 & 0.0 & 5.8 & 5.8
& 8.0 & 1.5 & 2.0 & 2.0 & 1.0 & 0.0 & 14.5 & 14.5 \\

\midrule

\rowcolor{benchHeader}
\multicolumn{2}{c}{\textbf{Configuration}}
& \multicolumn{8}{c@{\hspace{10pt}}}{\textbf{Unity}}
& \multicolumn{8}{c@{}}{\textbf{Roblox Studio}} \\
\cmidrule(lr){3-10}\cmidrule(l){11-18}

\rowcolor{white}
Scaffold & Model
& D1 & D2 & D3 & D4 & D5 & D6 & $\mathbf{F}$ & $\mathbf{O}$
& D1 & D2 & D3 & D4 & D5 & D6 & $\mathbf{F}$ & $\mathbf{O}$ \\
\midrule

\rowcolor{white}
\cellcolor{benchGroup}
& GPT-6 Astra
& 10.0 & 24.8 & 15.0 & 13.0 & 7.4 & 7.5
& \textbf{70.1} & \textbf{77.6}
& 10.0 & 25.5 & 18.0 & 12.5 & 6.0 & 2.5
& \textbf{72.0} & \textbf{74.5} \\

\rowcolor{benchStripe}
\cellcolor{benchGroup}
& GPT-5.6 Sol
& 9.2 & 23.4 & 13.2 & 11.8 & 6.0 & 5.4 & 63.6 & 69.0
& 9.4 & 11.1 & 6.1 & 7.6 & 4.3 & 4.3 & 38.6 & 42.9 \\

\rowcolor{white}
\cellcolor{benchGroup}
& GPT-5.6 Terra
& 9.3 & 13.6 & 8.3 & 5.7 & 4.2 & 2.6 & 41.1 & 43.7
& 6.5 & 4.4 & 2.1 & 3.2 & 2.2 & 1.3 & 18.4 & 19.7 \\

\rowcolor{benchStripe}
\cellcolor{benchGroup}\multirow{-4}{*}{Codex CLI}
& DeepSeek-V4-Pro
& 9.8 & 16.6 & 9.2 & 8.3 & 3.6 & 4.8 & 47.5 & 52.3
& 10.0 & 15.5 & 8.4 & 8.6 & 6.5 & 4.3 & 48.9 & 53.2 \\

\rowcolor{white}
DeepSeek Harness & DeepSeek-V4-Pro
& 10.0 & 14.6 & 11.6 & 8.8 & 3.1 & 5.3 & 48.1 & 53.4
& 9.5 & 7.5 & 5.0 & 2.5 & 2.5 & 0.8 & 27.0 & 27.8 \\

\rowcolor{benchStripe}
\cellcolor{benchGroup}Claude Code
& \claudemodel
& 9.6 & 22.1 & 12.7 & 11.3 & 4.8 & 5.2 & 60.5 & 65.7
& 9.0 & 9.0 & 3.0 & 10.0 & 4.5 & 3.8 & 35.5 & 39.3 \\

\rowcolor{white}
Kimi Code & \kimimodel
& 10.0 & 21.0 & 13.0 & 8.0 & 3.7 & 3.5 & 55.7 & 59.2
& 9.5 & 5.3 & 2.5 & 3.3 & 3.0 & 1.3 & 23.6 & 24.8 \\

\midrule

\rowcolor{benchHeader}
\multicolumn{2}{c}{\textbf{Configuration}}
& \multicolumn{8}{c@{\hspace{10pt}}}{\textbf{Unreal Engine}}
& \multicolumn{8}{c@{}}{\textbf{All engines}} \\
\cmidrule(lr){3-10}\cmidrule(l){11-18}

\rowcolor{white}
Scaffold & Model
& D1 & D2 & D3 & D4 & D5 & D6 & $\mathbf{F}$ & $\mathbf{O}$
& D1 & D2 & D3 & D4 & D5 & D6 & $\mathbf{F}$ & $\mathbf{O}$ \\
\midrule

\rowcolor{white}
\cellcolor{benchGroup}
& GPT-6 Astra
& 10.0 & 20.0 & 16.9 & 13.3 & 10.0 & 7.9
& \textbf{70.2} & \textbf{78.1}
& 9.9 & 21.3 & 13.9 & 13.1 & 6.8 & 6.5
& \textbf{65.0} & \textbf{71.4} \\

\rowcolor{benchStripe}
\cellcolor{benchGroup}
& GPT-5.6 Sol
& 9.4 & 5.6 & 3.7 & 6.8 & 3.4 & 2.7 & 28.8 & 31.5
& 9.5 & 12.6 & 6.9 & 8.0 & 4.0 & 4.3 & 41.0 & 45.3 \\

\rowcolor{white}
\cellcolor{benchGroup}
& GPT-5.6 Terra
& 8.5 & 2.1 & 3.5 & 4.6 & 1.4 & 1.4 & 20.0 & 21.4
& 8.1 & 5.8 & 4.7 & 4.9 & 2.3 & 2.3 & 25.8 & 28.1 \\

\rowcolor{benchStripe}
\cellcolor{benchGroup}\multirow{-4}{*}{Codex CLI}
& DeepSeek-V4-Pro
& 9.3 & 3.4 & 1.8 & 2.7 & 1.7 & 3.5 & 18.9 & 22.4
& 9.3 & 10.6 & 5.7 & 6.1 & 3.4 & 3.4 & 35.1 & 38.5 \\

\rowcolor{white}
DeepSeek Harness & DeepSeek-V4-Pro
& 10.0 & 9.0 & 7.2 & 10.0 & 2.6 & 4.0 & 38.7 & 42.7
& 9.9 & 10.8 & 7.3 & 7.4 & 2.8 & 3.5 & 38.2 & 41.6 \\

\rowcolor{benchStripe}
\cellcolor{benchGroup}Claude Code
& \claudemodel
& 9.0 & 3.8 & 4.7 & 6.0 & 1.7 & 0.9 & 25.1 & 25.9
& 9.0 & 10.4 & 6.2 & 8.5 & 4.1 & 3.3 & 38.3 & 41.5 \\

\rowcolor{white}
Kimi Code & \kimimodel
& 10.0 & 7.5 & 4.0 & 8.0 & 5.0 & 0.8 & 34.5 & 35.3
& 8.7 & 7.1 & 4.3 & 4.3 & 2.5 & 1.1 & 26.8 & 27.9 \\

\bottomrule
\end{tabular*}

\label{tab:main}
\label{tab:main-detailed}
\end{table}

\subsection{Buildable Is Not Playable}
\label{sec:res-gap}

Configurations reliably deliver projects that import, compile, and load
their entry scene: integration (D1) averages 8.9 to 9.7 of 10 points on
every engine (Table~\ref{tab:main}). Whether the game then works is
another matter: three in four cells earn under half of the 80 gameplay
points (D2--D5), and only GPT-6 Astra clears that bar on every engine.
The evaluator's player journey reaches the ending in only 16\% of runs,
with 4\% fully solved. Claude Code on Roblox
Studio, the median cell in gameplay, earns 9.0 of 10 integration points
but only a third of the gameplay points. Because integration barely
varies, the spread of $F$ is almost entirely a spread in gameplay; a
build check would rank most configurations as near-equals, whereas play
separates them by tens of points. Nor is the state interface the
obstacle: $\iface$ answers the handshake in 94\% of delivered projects.
Yet these contracts are within reach of the same technology under human
direction: the references that engineers built with coding agents are
all solved and average 87.3 of 90 on $F$ (96.4 of 100 on $O$),
against 65.0 for the best autonomous configuration and at most 41.0 for
the other six. The gap also varies by track: the share of gameplay points
runs from 23\% on Godot to 57\% on Unity, so the same build success
stands for very different amounts of playable game.

The scorecard also shows where games fail, and the answer shifts with
capability (Appendix Table~\ref{tab:weakest-dim}). We split the cells
into thirds by $F$ and find, in each, the gameplay dimension that is
weakest relative to its weight. In the lowest third it is most often
completeness (D2): much of the game is never built. In the middle third
it is mechanics (D3): the game exists, but its rules break under play.
In the top third it is lifecycle (D4), ahead of feedback (D5) and
mechanics: the game plays, but breaks across transitions, restarts, or
reopening, or fails to show the player what happened. Stronger agents
thus fail at later stages, a progression that a single pass/fail
outcome hides and that motivates replaying complete journeys and
checking visible responses during play.

\subsection{The Engine Is a First-Class Variable}
\label{sec:res-rank}
\label{sec:res-swap}

Every engine separates configurations widely, but not equally and not in
the same order. The spread of $F$ across configurations ranges from 29.0
points on Unity to 56.9 on Defold, and no two engines rank the
configurations alike: Kendall's $\tau$ between engines
ranges from 0.14 to 0.62 (Appendix Table~\ref{tab:rank-agreement}). Codex CLI with DeepSeek-V4-Pro, for instance,
ranks second of seven on Roblox Studio and last on Unreal Engine. GPT-6 Astra, the strongest
configuration, leads by 31.5 points on Unreal Engine and 23.1 on Roblox
Studio but by only 6.5 on Unity. An engine track can thus amplify a
capability difference that another hides, and a leaderboard built on one
engine inherits its bias.

Changing one factor at a time shows what transfers
(Appendix Table~\ref{tab:contrasts}). Within the GPT family the order holds on
every engine: GPT-6 Astra leads GPT-5.6 Sol by 6.5 to 41.4 points, and
Sol leads GPT-5.6 Terra by 5.1 to 22.5. Across families it does not: under Codex CLI,
DeepSeek-V4-Pro trails Sol by 26.7 points on Defold yet leads it by 12.7
on Godot. The scaffold can matter as much as the model: with
DeepSeek-V4-Pro fixed, moving from Codex CLI to DeepSeek Harness gains
19.8 points on Unreal Engine but loses 21.9 on Roblox Studio. A
single-engine leaderboard therefore ranks configurations for that engine
alone.

\Needspace{190pt}
\begin{wraptable}{r}{0.50\textwidth}
\providecolor{ogblue}{RGB}{31,119,180}
\providecolor{ogorange}{RGB}{230,120,30}
\providecolor{ogteal}{RGB}{26,150,130}
\captionsetup{font=footnotesize}
\caption{The same games on two engines: mean $F$ on the heavyweight
($\bar F_H$, rows) and script-centric ($\bar F_S$, columns) engine and
$\Delta F=\bar F_H-\bar F_S$, with standard deviations over games as
subscripts. Each pair pools its ten swapped games, five native to each
engine.}
\centering
\begingroup
\footnotesize
\setlength{\tabcolsep}{1.5pt}
\renewcommand{\arraystretch}{1.10}

\begin{tabularx}{\linewidth}{
  @{}cll*{3}{>{\centering\arraybackslash}X}@{}}
\toprule
\multicolumn{3}{c}{}
 & \multicolumn{3}{c}{
   \textbf{Script-centric engine ($S$)}} \\
\cmidrule(lr){4-6}

\rowcolor{ogblue!8}
 & \textbf{Engine} & \textbf{Metric}
 & \textbf{Defold} & \textbf{Godot} & \textbf{Roblox} \\
\midrule

\multirow{6}{*}{
  \rotatebox{90}{\footnotesize\textbf{Heavyweight ($H$)}}}
 & \multirow{3}{*}{\textbf{Unity}}
 & $\bar F_H$
 & \cellcolor{ogteal!22}
   $40.6_{\scriptscriptstyle\pm\,8.5}$
 & \cellcolor{ogteal!21}
   $38.0_{\scriptscriptstyle\pm\,8.9}$
 & \cellcolor{ogteal!19}
   $35.0_{\scriptscriptstyle\pm\,12.0}$ \\

 & & $\bar F_S$
 & \cellcolor{ogteal!16}
   $29.0_{\scriptscriptstyle\pm\,11.1}$
 & \cellcolor{ogteal!15}
   $27.6_{\scriptscriptstyle\pm\,6.4}$
 & \cellcolor{ogteal!21}
   $39.3_{\scriptscriptstyle\pm\,9.7}$ \\

 & & $\Delta F$
 & \cellcolor{ogblue!15}
   $\mathbf{+11.6}_{\scriptscriptstyle\pm\,4.3}$
 & \cellcolor{ogblue!14}
   $\mathbf{+10.4}_{\scriptscriptstyle\pm\,3.1}$
 & \cellcolor{ogorange!12}
   $\mathbf{-4.3}_{\scriptscriptstyle\pm\,5.4}$ \\

\cmidrule(lr){2-6}

 & \multirow{3}{*}{
   \makecell[l]{\textbf{Unreal}\\\textbf{Engine}}}
 & $\bar F_H$
 & \cellcolor{ogteal!27}
   $49.0_{\scriptscriptstyle\pm\,10.6}$
 & \cellcolor{ogteal!24}
   $45.0_{\scriptscriptstyle\pm\,12.1}$
 & \cellcolor{ogteal!18}
   $34.0_{\scriptscriptstyle\pm\,10.4}$ \\

 & & $\bar F_S$
 & \cellcolor{ogteal!16}
   $27.8_{\scriptscriptstyle\pm\,14.3}$
 & \cellcolor{ogteal!13}
   $23.5_{\scriptscriptstyle\pm\,13.2}$
 & \cellcolor{ogteal!14}
   $25.5_{\scriptscriptstyle\pm\,16.2}$ \\

 & & $\Delta F$
 & \cellcolor{ogblue!22}
   $\mathbf{+21.2}_{\scriptscriptstyle\pm\,11.5}$
 & \cellcolor{ogblue!23}
   $\mathbf{+21.5}_{\scriptscriptstyle\pm\,15.8}$
 & \cellcolor{ogblue!11}
   $\mathbf{+8.5}_{\scriptscriptstyle\pm\,7.6}$ \\

\bottomrule
\end{tabularx}

\label{tab:transfer-summary-mockup}
\endgroup
\end{wraptable}

\begin{wraptext}
\paragraph{Same game, different engine.}
The swap experiment holds the game fixed and changes only the engine:
for each of the six heavyweight--script-centric pairs, the five hardest
games of each engine are rebuilt on the other engine, which limits
ceiling effects on the heavyweight engines (Appendix~\ref{app:transfer}).
The heavyweight engine wins five of the six pairs
(Table~\ref{tab:transfer-summary-mockup}): Unity leads Defold and Godot
by 11.6 and 10.4 points, Unreal Engine leads Defold, Godot, and Roblox
Studio by 21.2, 21.5, and 8.5, and only Unity against Roblox Studio
reverses, by 4.3. Pooling both swap directions offsets any advantage a
game draws from its native engine, and with ten games per pair, each of
the five leads exceeds its standard deviation over games, hence more
than three standard errors.

The ordering matches how much an engine checks before play. Unity and
Unreal Engine compile their code, validate serialized scenes and
references, and ship test frameworks an agent can drive, so many errors
surface as diagnostics during construction; the advantage is widest
against Defold and Godot, whose dynamically typed scripts fail only at
run time, and narrowest against Roblox Studio, whose editor also
analyzes scripts. The native tracks agree. Kimi Code never plays or views its game
(Appendix Table~\ref{tab:trace-composition}): on Defold it gets no game
running, while on Unity it reaches $F=55.7$, a gap of 49.9 points against
6.5 to 21.7 for the other configurations across both tracks. An
agent that never runs its game depends entirely on checks before play.
Heavyweight engines are harder to drive but easier to check.

How much they help still varies by configuration: on one swapped game
the gain ranges from 7.0 to 77.9 points, and on another the two
scaffolds sharing DeepSeek-V4-Pro trade places (Appendix~\ref{app:transfer}).
Engine difficulty is therefore no fixed offset that could be calibrated
away.

\providecolor{ogblue}{RGB}{31,119,180}
\providecolor{ogorange}{RGB}{230,120,30}
\providecolor{ogteal}{RGB}{26,150,130}

\end{wraptext}

\subsection{Effort Does Not Buy Play}
\label{sec:res-cost}

Each construction trace shows what a game cost to build
(Figure~\ref{fig:cost}). The checks that favor heavyweight engines are
paid for in time: Unreal Engine runs reach the highest step and token
counts, averaging up to 295 workflow steps and 65M tokens per
configuration, because every build--play--repair iteration passes
through a long compile and import cycle. At the other end, Godot's
headless checks return short diagnostics, so the context resent with
every call grows slowly, and many Godot runs, especially early
stoppers, stay under 0.3M tokens.

\begin{figure}[!ht]
    \centering
    \includegraphics[width=\textwidth]{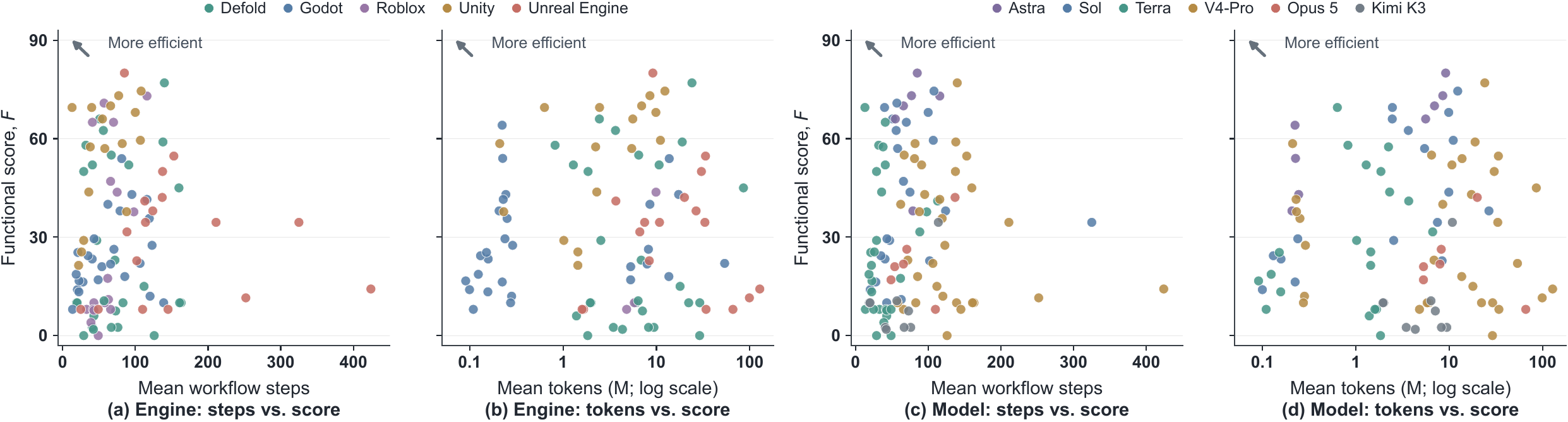}
    \caption{Construction cost against $F$ for a uniform sample of
game--configuration pairs: mean workflow steps (a, c) and mean tokens
on a log scale (b, d), colored by engine (a, b) or model (c, d;
DeepSeek-V4-Pro pools its two scaffolds).}
    \label{fig:cost}
\end{figure}

The time budget rarely binds. About nine in ten runs deliver before the
3600\,s limit, and those that reach it stall or loop, averaging only
16.3 points of $F$. Most runs thus end when the agent judges its game
finished, so the 6--14 engineer-hours behind each reference do not
explain the gap. That judgment is rarely tested by play: playing and
viewing the game take at most 13.3\% of any configuration's steps. Even
at that maximum, GPT-6 Astra and DeepSeek-V4-Pro under Codex CLI differ
by 29.9 points, so play frequency alone does not decide $F$.
Nor do steps or tokens track $F$: games of similar cost score anywhere
from near zero to 80, and no game whose runs average more than 200 steps
reaches 40 points. The best games sit toward the upper left, where
GPT-6 Astra reaches its top scores on moderate budgets, so capability
shows up as efficiency, not persistence.

\FloatBarrier
\Needspace{200pt}
\subsection{Effect of Game Completion Boost}
\label{sec:res-method}

\begin{wraptable}{r}{0.50\textwidth}
\captionsetup{font=footnotesize}
\caption{GCB under Codex CLI on the 15 games, three per engine
track, ordered by baseline $O$.
$\Delta F$ and $\Delta O$ are gains over the paired baseline.}
\centering
\begingroup
\footnotesize
\setlength{\tabcolsep}{2.8pt}
\renewcommand{\arraystretch}{1.12}
\begin{tabularx}{\linewidth}{@{}Yrrrr@{}}
\toprule
\rowcolor{ogblue!8}
\textbf{Configuration} & $\boldsymbol{F}$ & $\boldsymbol{\Delta F}$ & $\boldsymbol{O}$ & $\boldsymbol{\Delta O}$ \\
\midrule
DeepSeek-V4-Pro & \cellcolor{ogteal!16}50.6 & --- & \cellcolor{ogteal!17}54.8 & --- \\
DeepSeek-V4-Pro + GCB & \cellcolor{ogteal!25}\textbf{71.6}
  & \cellcolor{ogblue!18}\textbf{+21.0}
  & \cellcolor{ogteal!27}\textbf{76.3} & \cellcolor{ogblue!18}\textbf{+21.5} \\
\cmidrule(lr){1-5}
GPT-5.6 Sol & \cellcolor{ogteal!10}27.0 & --- & \cellcolor{ogteal!11}30.9 & --- \\
GPT-5.6 Sol + GCB & \cellcolor{ogteal!15}\textbf{41.1}
  & \cellcolor{ogblue!13}\textbf{+14.1}
  & \cellcolor{ogteal!16}\textbf{45.6} & \cellcolor{ogblue!13}\textbf{+14.7} \\
\cmidrule(lr){1-5}
GPT-5.6 Terra & \cellcolor{ogteal!5}9.7 & --- & \cellcolor{ogteal!6}10.5 & --- \\
GPT-5.6 Terra + GCB & \cellcolor{ogteal!12}\textbf{34.0}
  & \cellcolor{ogblue!22}\textbf{+24.3}
  & \cellcolor{ogteal!13}\textbf{36.3} & \cellcolor{ogblue!22}\textbf{+25.8} \\
\bottomrule
\end{tabularx}
\endgroup

\label{tab:completion-skill}
\end{wraptable}

\begin{wraptext}
We evaluate GCB under Codex CLI with three models on 15 high-difficulty
games, three per engine track, fixed before any GCB run
(Appendices~\ref{app:tasks} and~\ref{app:recovery}). Each paired
comparison holds the task, native prompt, engine-access and image tools,
and budget fixed, and the baseline is the strongest of a game's three
native runs by $O$, so it sits at or above the game's three-run mean
used in Table~\ref{tab:main}.

GCB improves $F$ and $O$ for every model
(Table~\ref{tab:completion-skill}), raising mean $F$ by
14.1--24.3 points and mean $O$ by 14.7--25.8; Terra more than triples
its $F$. Each gains most in the bottleneck that
Section~\ref{sec:res-gap} finds at its score level. DeepSeek-V4-Pro,
the strongest baseline here, gains most in lifecycle
(D4, $+8.6$), while Sol and Terra gain most in completeness
and mechanics (D2/D3, $+5.3/+4.3$ and $+8.0/+5.3$). D6 gains 0.5--1.5 points and D1 changes by $-0.7$ to $+3.3$,
so most of the gain is gameplay.

\end{wraptext}

\FloatBarrier
\section{Conclusion}
\label{sec:conclusion}

\bench{} evaluates complete-game creation on five production engines
through real-input play. Agents usually deliver projects that build yet earn under half the
gameplay points, and most stop before the time limit, whereas engineers
directing coding agents reach near the maximum. No two engines rank configurations alike, and the same game
usually scores higher on the heavyweight engines, which are harder to
drive but easier to check. GCB turns these findings into a
playtest-and-repair workflow that improves every evaluated model. For
evaluation practice, the results favor scoring games by play, reporting
each engine separately, and logging build cost. The shared evaluation protocol and execution traces support comparisons of functional performance across engine tracks and agent configurations. Game
creation thus tests whether an agent can inspect and repair its own work
before declaring it done, a capability needed to build simulated worlds
from specifications.

\section*{Limitations}
\label{sec:limitations}
\bench{} evaluates single-player games built offline from a written
design document. It measures whether a game meets its specification and
leaves player enjoyment, online multiplayer, live services, and design
without a specification outside its scope. The
five engines span script-centric and heavyweight production engines;
other engines and web frameworks can join as new tracks under the same
contract language. Engine and editor versions are frozen, so newer
releases enter as new benchmark versions rather than shifting existing
scores. GCB is evaluated under one scaffold with three models, excluding
GPT-6 Astra, on high-difficulty games from each track; its effect under
other scaffolds and on the strongest model remains to be measured. The
visual judge shares its model with one evaluated configuration, so
configurations are compared on $F$, which excludes D6. Built by
engineers working with coding agents, the references mark what
human-directed agents reach rather than unaided human skill. A swap
changes the starter and tool manual with the engine, so its effect
covers the whole toolchain, not the engine core alone. Finally, the
heavyweight tracks need full editors and build toolchains, and one
Unreal Engine run can consume tens of millions of tokens, so a complete
evaluation is costlier than a browser-based benchmark. This cost also
bounds the scale: with ten tasks per engine, seven configurations, and
swaps on the five hardest games of each track, engine-level rankings and
swap effects carry wide spreads, which we report.

\section*{Ethical Considerations}
All tasks are adapted from permissively licensed open-source games or
written from scratch; adapted games are renamed and re-parameterized,
and bundled art and audio are used under their licenses with
attribution. The six annotators and engineers who built the tasks were
compensated in accordance with applicable labor laws and regulations. Every engine runs locally and no game is published online,
so the benchmark involves no personal or player data. As with other
code-generating agents, game-creation agents could be misused to clone
commercial games; the release contains only newly written documents, licensed assets, and our own evaluator and run traces, and no material from commercial games.

\FloatBarrier
\bibliographystyle{assets/plainnat}
\bibliography{omnigame_bench_draft}

\clearpage
\beginappendix

\section{Engine Tracks and Toolchains}
\label{app:engines}

Each track is pinned to the engine build of Table~\ref{tab:engines}, one
editor build, and one command-line entry point, and a run that resolves
any other version is discarded. Unity tasks use the 2D and 3D core
packages and no Asset Store content; Unreal Engine tasks use its C++
toolchain and also permit Blueprint; Godot tasks use GDScript only; and
Defold tasks use the stock build server. Roblox Studio submissions are
\texttt{.rbxl} place files play-tested in local Studio sessions and
never published to a live service. Starters bundle CC0 or CC-BY art and
audio, adapted source games are licensed under MIT, Apache-2.0, CC0, or
zlib, and per-task attribution ships with the release.

\paragraph{Checks before play.}
The tracks differ in how much an engine checks a game before it runs,
and the tool manual exposes each engine's own checks to the agent
(Table~\ref{tab:engines}, fifth column). On Unity and Unreal Engine, C\#
and C++ compilation rejects type and reference errors, the editor
validates serialized scenes and assets, and built-in test frameworks run
scripted checks inside the engine. Roblox Studio analyzes Luau scripts in
the editor. Godot parses GDScript on load, and Defold validates resources
and Lua syntax at build time; beyond that, errors in their dynamically
typed scripts surface only when the game runs. These checks are the
diagnostics an agent can act on before it plays its own game.

\paragraph{Starter and tool manual.}
Each starter $s_0$ opens, builds, and runs on its locked version but
implements none of the game. It provides an empty entry scene, the
task's licensed art and audio, naming tables for scenes, entry points,
and state keys, and the public smoke checks of Appendix~\ref{app:gdd}.
The tool manual $M$ documents command-line build and headless run,
editor automation, import and packaging, the checks above, log
locations, and tools for capturing and inspecting a running game.
Starters and manuals follow a common section structure across tracks
(Appendix~\ref{app:fairness}).

\begin{table}[H]
\caption{Engine builds, project representations, pre-play diagnostics,
and evaluator adapters for the five tracks.}
\centering
\begingroup
\footnotesize
\setlength{\tabcolsep}{3.4pt}
\renewcommand{\arraystretch}{1.16}

\begin{tabularx}{\textwidth}{
  @{}
  P{0.115\textwidth}
  P{0.13\textwidth}
  P{0.085\textwidth}
  Y Y
  P{0.125\textwidth}
  @{}
}
\toprule
\textbf{Engine}
& \textbf{Engine build}
& \textbf{Language}
& \textbf{Scene representation}
& \textbf{Checks before play}
& \textbf{Evaluator adapter} \\
\midrule

\multicolumn{6}{@{}l}{\textit{Script-centric engines}} \\

\rowcolor{black!3}
Defold
& \texttt{1.13.0}
& Lua
& Data-driven game objects and collections.
& Resource and reference validation; Lua syntax checks during build.
& Engine Lua hook. \\

Godot
& \texttt{4.7.1.stable}
& GDScript
& Node trees serialized as text scenes.
& GDScript parsing and scene-loading diagnostics.
& Headless script. \\

\rowcolor{black!3}
Roblox Studio
& \texttt{0.742.0.\allowbreak7421053}
& Luau
& Studio instance trees stored in local \texttt{.rbxl} files.
& Luau analysis and editor diagnostics.
& Studio plugin and local play-test. \\

\midrule
\multicolumn{6}{@{}l}{\textit{Heavyweight engines}} \\

Unity
& \makecell[l]{6.0 LTS\\\texttt{6000.0.80f1}}
& C\#
& YAML scenes and prefabs with an asset-import pipeline.
& C\# compilation, serialized-reference checks, and Unity tests.
& PlayMode harness. \\

\rowcolor{black!3}
Unreal Engine
& \makecell[l]{\texttt{5.7.4}\\CL~51494982}
& \makecell[l]{C++\\Blueprint}
& Binary assets, C++ modules, and build targets.
& C++ and Blueprint compilation, asset validation, and Automation tests.
& Automation and functional tests. \\

\bottomrule
\end{tabularx}

\label{tab:engines}
\endgroup
\end{table}

\section{Cross-Engine Comparability}
\label{app:fairness}

A Defold score and an Unreal Engine score are comparable only if the
scale, not the difficulty, is held fixed. Five controls fix the scale and
leave the engine as the variable.
\begin{enumerate}[leftmargin=*,itemsep=1pt,topsep=2pt]
\item \textbf{One contract language.} Every track instantiates the same
contract symbols and six dimensions, and a task keeps its leaf families
on every engine; only the evidence route through the engine adapter
differs.
\item \textbf{Fixed family budgets.} Within a dimension, budgets
$\beta_{d,f}$ allocate weight to families and stay fixed across engines,
so an obligation weighs the same whether it takes twelve leaves on
Unreal Engine or three on Defold.
\item \textbf{Matched intended difficulty.} Every track has the same
3/5/2 easy/medium/hard composition under the four-axis annotation of
Appendix~\ref{app:tasks}.
\item \textbf{Matched ceilings.} A task enters any track only if its
reference implementation scores at least 90 on $O$, and the released
references score at least 92.5 on every track
(Appendix~\ref{app:construction}), so the full scale is reachable
everywhere and a low track mean reflects the agents.
\item \textbf{Matched inputs and budgets.} Starter structure, GDD
structure, smoke checks, tool-manual sections, and the 3600\,s budget
are identical across tracks, and the task prompt
(Appendix~\ref{app:prompts}) contains no engine-specific hints.
\end{enumerate}
The controls preserve each engine's scene model, asset pipeline, build cycle, and checks before play
(Table~\ref{tab:engines}). These set both what each build--play--repair
iteration costs and how many errors surface before the game runs, and
they are what the benchmark measures. Since tracks have their own games, results are reported per track, and the cross-group swaps of
Appendix~\ref{app:transfer} isolate the engine on a fixed game.

\section{Game Design Document and State Interface}
\label{app:gdd}

This appendix details the two public documents of
Section~\ref{sec:task-formulation} that define a task for the agent:
the GDD and the state interface it requires.

\paragraph{Document structure.}
Every task uses the same fifteen GDD sections to make the specification's scope explicit. Table~\ref{tab:gdd} maps each
section to the contract symbol it populates and to the dimensions its
leaf checks score.

\begin{table}[H]
\caption{The shared GDD structure and its mapping to contract symbols
and evaluation dimensions.}
\centering
\begingroup
\footnotesize
\setlength{\tabcolsep}{3.4pt}
\renewcommand{\arraystretch}{1.16}
\rowcolors{2}{LinkPanel}{white}

\begin{tabularx}{\textwidth}{
  @{}r P{0.24\textwidth} Y c P{0.105\textwidth}@{}
}
\toprule
\textbf{\#}
& \textbf{GDD section}
& \textbf{Information provided to the agent}
& \textbf{Symbol}
& \textbf{Dimensions} \\
\midrule

1
& Pitch and core loop
& Game premise, session structure, and target duration.
& ---
& D2 \\

2
& Engine and toolchain
& Engine build, dependencies, and build/run commands.
& ---
& D1 \\

3
& Controls
& Input devices, bindings, actions, and press/hold/release behavior.
& $\mathcal{A}$
& D3, D5 \\

4
& Entities and attributes
& Required actors, props, UI elements, and numerical attributes.
& $\mathcal{N}$
& D2, D3 \\

5
& Scene inventory
& Entry scene, levels, scene contents, and connections.
& $\mathcal{S}_a$
& D1, D2 \\

6
& Mechanics and rules
& Action resolution, costs, resources, and collision behavior.
& $\mathcal{S}_r$
& D3 \\

7
& Temporal rules
& Cooldowns, timing windows, durations, and event ordering.
& $\mathcal{T}$
& D3 \\

8
& Branches and endings
& Success, failure, rejection paths, and terminal states.
& $\mathcal{B}$
& D3, D4 \\

9
& Progression
& Unlock order, difficulty progression, and scoring rules.
& $\mathcal{B}$
& D2, D4 \\

10
& Invariants
& Conditions that must remain true throughout play.
& $\mathcal{I}$
& D3, D4 \\

11
& HUD and feedback
& Readouts, prompts, and audiovisual responses to player actions.
& ---
& D5 \\

12
& Persistence and lifecycle
& Pause, restart, save, and fresh-process reopening behavior.
& $\mathcal{P}$
& D4 \\

13
& Asset manifest
& Permitted assets, naming conventions, and prohibited sources.
& ---
& D2, D6 \\

14
& State interface
& Required keys, types, semantics, and interface commands.
& $\iface$
& D1, D3, D4 \\

15
& Acceptance checklist
& Public smoke checks for basic delivery and integration.
& ---
& D1 \\

\bottomrule
\end{tabularx}

\label{tab:gdd}
\endgroup
\end{table}

\paragraph{From requirement to leaf.}
Each sentence of the normative sections becomes one or more leaf checks,
written against the contract rather than any implementation.
Table~\ref{tab:gdd-example} traces an excerpt of \emph{Driftfolk
Chronicle}. A requirement is checked on its positive branch and, where
the GDD implies one, on its rejected branch: the dash is tested both when
it should fire and when it should not. Every assertion reads only
declared $\iface$ keys, HUD elements, logs, or frames, so the same rows
apply to any implementation on any engine, and only the adapter that
sends the input and collects the evidence changes.

\begin{table}[H]
\caption{Example requirement-to-check mapping. Sym., Dim., and $w$
denote the contract symbol, score dimension, and leaf weight
(1 minor, 2 major, 3 critical).}
\centering
\footnotesize
\setlength{\tabcolsep}{3.4pt}
\renewcommand{\arraystretch}{1.16}
\rowcolors{2}{LinkPanel}{white}

\begin{tabularx}{\textwidth}{
  @{}r P{0.25\textwidth} c c c P{0.105\textwidth} Y@{}
}
\toprule
\textbf{GDD\,\S} & \textbf{Requirement}
& \textbf{Sym.} & \textbf{Dim.} & $\boldsymbol{w}$
& \textbf{Route} & \textbf{Leaf assertion} \\
\midrule

5
& The game opens on the title menu; \emph{Start} enters the Harbor level.
& $\mathcal{S}_a$ & D1 & 2
& Static, input
& The entry scene is the title menu; after \emph{Start},
  \texttt{level.id} is \texttt{harbor}. \\

3
& Press Shift to dash in the facing direction.
& $\mathcal{A}$ & D3 & 3
& Input + read
& After Shift, \texttt{player.state} is \texttt{dash}
  and \texttt{player.pos} moves along \texttt{player.facing}. \\

6
& A dash spends one charge.
& $\mathcal{S}_r$ & D3 & 3
& Input + read
& \texttt{player.charges} decreases by exactly one per dash. \\

7
& A new dash is rejected for 0.4\,s after the last one.
& $\mathcal{T}$ & D3 & 2
& Input + read
& A second dash at 0.2\,s is rejected;
  one at 0.5\,s is accepted. \\

8
& With no charges left, dash does nothing.
& $\mathcal{B}$ & D3 & 2
& Input + read
& At \texttt{player.charges}$=0$, Shift leaves
  position and charges unchanged. \\

10
& Charges never exceed the maximum.
& $\mathcal{I}$ & D3 & 2
& Input + read
& During a 10\,s recharge, \texttt{player.charges}
  $\leq$ \texttt{player.max\_charges} at every sample. \\

11
& The charge meter always shows the remaining charges.
& --- & D5 & 1
& Input + frame
& After each dash, the HUD meter matches
  \texttt{player.charges} within 0.2\,s. \\

12
& Charge upgrades survive quitting the game.
& $\mathcal{P}$ & D4 & 3
& Lifecycle
& After reopening in a fresh process,
  \texttt{player.max\_charges} equals its value before quitting. \\

13
& Characters use the bundled sprite set.
& --- & D6 & 1
& Visual
& Asset assembly is rated against the reference capture at each registered state. \\

\bottomrule
\end{tabularx}

\label{tab:gdd-example}
\end{table}

\paragraph{State interface.}
$\iface$ has one read accessor, \texttt{OmniState.Get(key)}, over the
task's declared keys and two commands, \texttt{OmniFlow.LoadState(id)}
and \texttt{OmniFlow.Restart()}. Keys are typed and documented in GDD
section~14; their identifiers are fixed by the naming table, and their
implementation is free. Four rules govern its use.
(i)~$\iface$ never performs a player action; every action arrives as a
real input event.
(ii)~$\iface$ enters only states the game itself can reach, through the game’s own loading path, and state preparation ends before the
first input of a probe.
(iii)~Every reading used for scoring is triangulated against at least
one independent route: a frame capture, a log line, or a second key it
constrains. A disagreement resolves to \textsc{observer\_failed} on its
first occurrence and, if it recurs for the same key, to
\textsc{fail\_observed} on the interface leaf.
(iv)~$\iface$ is a declared precondition of every leaf that reads it. If
the handshake fails or a required key is missing or mistyped, those
leaves resolve to \textsc{product\_blocked} and count as failures, while
leaves that do not read $\iface$, such as static checks, are still
judged.
Rules (iii) and (iv) close two shortcuts. An accessor returning
constants fails its own leaf and earns no behavioral leaves, and a
missing accessor forfeits every leaf that reads it.

\paragraph{What $\iface$ can and cannot certify.}
Reporting state truthfully is part of the deliverable and far simpler
than building the game it describes, so a game whose $\iface$ misreports
loses the interface leaf and cannot pass any leaf that reads the false
key. $\iface$ cannot earn credit through self-reported state alone. First, it
has no path to outcomes: after state preparation, every outcome a probe
checks must follow from real input injected at the engine's input
layer. Second, its readings must hold across probes the submission
cannot anticipate, including rejected actions, timing windows,
invariants sampled over seconds, and closed-loop journeys; a key that
stays correct across such inputs, or two keys that stay consistent,
must be computed from play by the game's rules, which is what D3
scores. Third, what a correct key cannot show, namely that the world
and the screen enact the state, rests on evidence the engine produces:
rendered frames, the engine's UI tree and its audio and animation
events (Appendix~\ref{app:stages}), a fresh process for reopening, and
reference captures for D6. A first disagreement with such evidence is
charged to the evaluator, since capture timing alone can cause one, and
removes the leaf from the denominator; a recurrence fails the interface
leaf. All exclusions together stay within 2.9\% of leaf weight on every
engine, and a dimension whose judged weight falls below 0.90 is reported
as inconclusive rather than scored (Appendix~\ref{app:scoring}). The
submissions bear this out: $\iface$ answers its handshake in 94\% of
projects, yet only 16\% of runs complete the journey, and the
\emph{Captain2D Platformer} run of Appendix~\ref{app:results} publishes
fresh, well-formed state yet scores $O=15.2$.

\paragraph{Public smoke checks.}
Each task carries two to four public smoke checks, always including that
the project imports and builds without error, that the entry scene loads
and enters play mode, and that $\iface$ answers a handshake query with
the declared key set. They carry no score and let an agent confirm that
its game runs before it stops.

\section{Evaluator Stages and Engine Adapters}
\label{app:stages}
\label{sec:evaluation}

The evaluator accepts a game only by playing it
(Section~\ref{sec:scoring}). This appendix describes how it models the
game, plays it from entry to ending, and collects evidence on each
engine.

\paragraph{State graph.}
The evaluator models a game as a finite state graph. Its nodes are the
menu or entry, local level states, pause, death, success, restart, and a
clean reopen; its edges are real player actions or lifecycle
transitions. Three units are tested: behavior within one local state, a
single transition, and a bounded chain of adjacent transitions. To reach
a later level, a private fixture prepares a valid local state through
the game's own loading and initialization paths. Preparation ends
before the first player action; from then on the evaluator sends only
normal input and checks each obligation in full, including its negative
branches and cleanup.

\paragraph{Player journey.}
The journey plays the whole game from the entry scene to the terminal
state with public controls only and never calls
\texttt{OmniFlow.LoadState}. Because $D$ fixes levels, rules, and
constants, the reference path is written as a sequence of
contract-level actions, such as reaching an exit or clearing a room,
rather than as raw key timings. The evaluator executes each action as a
closed-loop input policy: it issues real input, steers toward targets
whose positions it reads through $\iface$, and polls $\iface$ until the
action's postcondition holds. Implementations with different pacing,
animation timing, or scene organization therefore complete the same
path. The journey budget is twice the number of input steps on the
reference path. A journey that exceeds the budget or stalls on an action leaves $J$ unset.

\paragraph{Stages.}
Table~\ref{tab:stages} lists the six evaluator stages, the evidence each
collects, and the dimensions or outcome flags it feeds.

\begin{table}[H]
\caption{Evaluator stages, their evidence sources, and the dimensions or
outcome flags they feed.}
\centering
\small
\begin{tabularx}{\linewidth}{@{}P{0.21\linewidth}YP{0.13\linewidth}@{}}
\toprule
Stage & Evidence & Feeds \\
\midrule
Static          & import and compile, entry scene, serialized references, manifests & D1, D2 \\
Local dynamic   & real input, $\iface$ snapshots, collision, AI, and resource assertions, negative controls & D2, D3, D5 \\
Transition      & scene change, resource transfer, death and restart, pause and resume, terminal freeze & D3, D4 \\
Player journey  & entry-to-terminal play with public controls, closed-loop input, step budget & $J$ \\
Fresh lifecycle & dirty close, new process, clean entry, restored references & D4, $\Lambda$ \\
Visual          & reference-state captures, paired reference images, written rubric, artifact hashes & D6 \\
\bottomrule
\end{tabularx}
\normalsize

\label{tab:stages}
\end{table}

\paragraph{Adapters.}
One semantic protocol compiles to five engine adapters: a Unity PlayMode
harness, an Unreal Engine automation and functional-test entry point, a
Godot headless script, a Defold engine Lua hook, and a Roblox Studio
plugin that drives local play-test sessions. Each adapter injects input
inside the engine process at its input layer rather than synthesizing
operating-system events, which keeps input timing reproducible and
removes window-focus instability. State reads go through $\iface$, so no
adapter depends on the submission's class layout.

\paragraph{Feedback evidence.}
Each D5 leaf names a player action, the response that GDD section~11
declares for it, and the channel that must carry the response. After
sending the action, the evaluator checks that channel within a fixed
window: a HUD element in the engine's UI tree whose value matches the
corresponding $\iface$ reading, an audio or animation event in the
engine log, or a pixel change in the declared screen region. Each check
is a deterministic comparison, so feedback is scored by execution, not
by model judgment.

\paragraph{Visual protocol.}
D6 compares the submission with its reference game state by state. A
task registers the reference states its game can show, such as the
menu, early and later play, a management screen, success, and failure;
each registered state is one D6 leaf, 3.7 per task on average
(Table~\ref{tab:rubric}). At each state, a vision--language model
(\texttt{gpt-5.6-sol}, reasoning effort \texttt{xhigh}) at fixed
decoding receives the reference capture, the submission's capture, and
the state label, and rates five items against written anchors
(Listing~\ref{lst:judge-prompt}): scene completeness and asset
assembly, composition and layout, scale and layering, style and color
coherence, and overall finish, each from 0 to 2 in half-point steps.
The judge is told to ignore gameplay, code, and internal state and to
tolerate differences in capture timing, so D6 scores presentation
alone. Three samples per state are aggregated by the median, and the
leaf takes the median total, divided by 10, as its credit. On a
120-image audited subset, model and human ratings agree at Spearman
$\rho=0.81$, and rescoring the same artifact moves D6 by at most 1.5
points. Because the judge model also builds games as one of the
evaluated configurations, D6 enters only $O$, and configurations are
compared on $F$ (Section~\ref{sec:setup}).

\section{Scoring Algorithm and Leaf States}
\label{app:scoring}

Section~\ref{sec:scoring} defines the scorecard; this appendix specifies
how each leaf is resolved and aggregated.

\paragraph{Leaf schema.}
A leaf is a tuple $\ell=(\texttt{id},d,f,w,r,\mathrm{pre})$ with
dimension $d\in\{1,\dots,6\}$ (Table~\ref{tab:rubric}), family $f$,
weight class $w\in\{1,2,3\}$ (minor, major, critical), a registered
evidence route $r$ (static, input-and-read, lifecycle, or visual), and
preconditions $\mathrm{pre}$ over other leaves. Leaf identifiers and
family budgets $\beta_{d,f}$ are frozen with the task before any model
is run.

\begin{table}[H]
\caption{The six outcome dimensions, shared by all five tracks. $W_d$ is
the dimension weight and Leaves the mean number of leaf checks per task
(47.2 in total). D1--D5 are scored by execution and D6 by a
vision--language model.}
\centering
\footnotesize
\setlength{\tabcolsep}{2.5pt}
\begin{tabular}{@{}l r r l P{0.47\columnwidth}@{}}
\toprule
ID & $W_d$ & Leaves & Name & Primary evidence \\
\midrule
D1 & 10 & 5.1  & Integration  & import, compile, entry scene, serialized references, manifests \\
D2 & 30 & 14.0 & Completeness & required entities, levels, modes, UI, authored content \\
D3 & 20 & 11.3 & Mechanics    & actions, rules, resources, collision, AI, negative branches \\
D4 & 20 & 8.8  & Lifecycle    & transitions, terminal state, restart, save, fresh reopen \\
D5 & 10 & 4.3  & Feedback     & controls, HUD, observable response to every player action \\
D6 & 10 & 3.7  & Presentation & asset assembly, layout, layering, style, and finish against reference captures \\
\bottomrule
\end{tabular}
\normalsize

\label{tab:rubric}
\end{table}

\paragraph{Procedure.}
The evaluator resolves the leaves of a task in order, and each leaf ends
in one of the six states of Table~\ref{tab:leaf}.
\begin{enumerate}[leftmargin=*,itemsep=1pt,topsep=2pt]
\item \emph{Preconditions.} An unmet precondition makes the leaf
\textsc{product\_blocked} when the submission causes the block and
\textsc{setup\_failed} when the fixture does.
\item \emph{Evidence.} The evaluator collects evidence along the leaf's
registered route and, if the evidence is missing or self-inconsistent,
retries once through an alternate route.
\item \emph{Judgment.} Decisive evidence yields \textsc{pass} or
\textsc{fail\_observed}; otherwise the leaf becomes
\textsc{observer\_failed} or \textsc{inconclusive}. A D6 leaf with a
valid rating is judged and takes that rating, divided by 10, as its
credit.
\item \emph{Coverage.} The coverage of a dimension is its judged weight
divided by its total weight. A dimension below 0.90 is re-run once and,
if still below, reported as inconclusive.
\item \emph{Aggregation.} Family pass ratios, dimension scores, $F$, and
$O$ follow Section~\ref{sec:scoring}.
\end{enumerate}

\begin{table}[H]
\centering
\begin{minipage}{0.90\textwidth}

\caption{Leaf states and their effect on the score. The first three are
judgments of the submission; the last three leave the denominator.}
\centering
\footnotesize
\setlength{\tabcolsep}{3pt}
\begin{tabularx}{\linewidth}{@{}llY@{}}
\toprule
State & Score effect & Meaning \\
\midrule
\textsc{pass}             & full weight          & obligation observed under valid preconditions \\
\textsc{fail\_observed}   & zero, in denominator & submission violates the obligation under valid observation \\
\textsc{product\_blocked} & zero, in denominator & submission itself prevents the required state or action \\
\textsc{setup\_failed}    & excluded             & fixture preparation failed, also through the alternate route \\
\textsc{observer\_failed} & excluded             & evidence acquisition failed or disagreed across routes \\
\textsc{inconclusive}     & excluded             & evidence insufficient; never inferred from absence \\
\bottomrule
\end{tabularx}
\normalsize

\label{tab:leaf}
\end{minipage}
\end{table}

\paragraph{Propagation.}
\textsc{product\_blocked} propagates only along declared preconditions
and only within the same chain of states: a level that never loads blocks that level's mechanics leaves, but not the menu leaves or another level
reachable through $\iface$. Non-judgment states never propagate. Because
the evaluator can enter declared states directly, an early break costs
its own subtree rather than the rest of the task.

\paragraph{Outcome flags.}
Each run also yields three binary outcomes that describe the game as a
whole. The journey flag $J$ is set when the evaluator completes the
player journey of Appendix~\ref{app:stages} within its budget; the
lifecycle flag $\Lambda$ is set when restart and a fresh-process reopen
preserve every requirement in $\mathcal{P}$; and a run is \emph{solved}
when $J$ and $\Lambda$ are set and every critical leaf passes. The
flags are stricter than the scorecard: a game can earn most of its
leaves in states entered through $\iface$ and still leave $J$ unset
because one transition on the public path is broken. Across native runs, $J$ is set in only 16\% of runs and 4\% are solved,
whereas projects earn 92\% of integration points on average.
\section{Construction, Calibration, and Leakage Control}
\label{app:construction}

This appendix details the authoring, calibration, and leakage control of
the tasks in Section~\ref{sec:construction}.

\paragraph{Authoring.}
For an adapted task, two annotators independently play the source game
to completion and write the GDD from player-visible behavior, reading
the implementation only afterwards for parameter values. For an
original task, one annotator writes the GDD from a design brief, and a
second reviews it against the fifteen-section template, flagging any
requirement that the player cannot observe or that admits two readings.
In both cases a third annotator adjudicates disagreements and owns the
frozen contract. The contract is frozen before the evaluator is
written, and the evaluator author sees $C$ but not $G$.

\paragraph{Annotators and engineers.}
Six annotators and engineers produced the tasks under the separation of
roles described above. They took part and were compensated in
accordance with applicable labor laws and regulations.

\paragraph{Calibration suite.}
Each task is calibrated on four kinds of artifact, and each kind tests
a different property of the evaluator $V$. The reference implementation
$G$ shows that the full scale is reachable. An equivalent
implementation $G'$, written from $C$ by a different engineer with a
deliberately different structure, shows that $V$ rewards behavior
rather than one way of building it. The untouched starter $s_0$ shows
that a project which builds but implements none of the game earns only
a few integration points. Four mutants of $G$, which break compilation,
feedback, a transition, or persistence, show that $V$ catches a fault
where it occurs and nowhere else. Admission requires, on $O$, $G$ to
score at least 90, $G'$ to land within 2.5 points of $G$, $s_0$ to score
at most 6, and every mutant to be detected with an off-target drift of
at most 2.0 points. A mutant's target is its expected dependency subgraph,
derived from the frozen leaf preconditions before the mutant is
evaluated: the leaves it breaks, every leaf whose declared preconditions
reach them, and, for a fault on the public path, the journey flag $J$
together with the presentation cap it controls. Detection requires a
failure inside this subgraph, and drift is the change in score
contributed by the leaves outside it. The four mutant classes therefore
test different properties. A compilation mutant blocks every leaf that
needs a running game, so its subgraph covers most of the task and only
its detection is at stake. A feedback or persistence mutant has a
subgraph confined to its D5 or D4 leaves. A transition mutant breaks
the journey and its own transition leaves, while the next level, still
entered through $\iface$, keeps its mechanics leaves; it therefore
tests the locality on which Section~\ref{sec:scoring} relies. A task
that misses any threshold is re-authored from its contract and
calibrated again, and several tasks went through this loop before
release. Every released task meets these criteria on every
track: references average 96.4 and none scores below 92.5, equivalent
implementations land 2.0 points from their references on average and
never more than 2.5, untouched starters score 4.1, and all 200 mutants, four per task, are detected with an off-target
drift of at most 2.0 points. Three evaluations of one artifact differ by at most
1.8 points, and their scripted leaves agree at $\kappa=0.94$.

\paragraph{Release audit.}
Before release, the input hash of $s_0$, $D$, and $M$, the evaluator
revision, the engine lock, and the evidence manifest are bound into one
record, and every calibration check is re-run on a clean import of the
frozen starter.

\paragraph{Leakage control.}
Leakage is controlled at three layers. At the \emph{artifact} layer,
starters are rebuilt from scratch, carry no version-control history,
and share no content hash with any public repository. At the
\emph{specification} layer, the GDD, state interface, naming tables,
and tool manual are newly written, and adapted games are renamed and
re-parameterized. At the \emph{evaluation} layer, the evaluator, leaf
schemas, and fixtures are hidden from the agent during a run, and
reference implementations, equivalent implementations, and their
mutants are never released; every released file carries a canary string
so that later training corpora can be screened for the benchmark. A
memorization probe tests the result: without the starter, each
evaluated model is asked to reproduce the file names, scene names, and
constants of a task, and its answers are matched against the task's
artifacts. Agreement stays at the chance level obtained by matching
each task against the names and constants of the other tasks.

\section{Task Inventory and Difficulty Annotation}
\label{app:tasks}

This appendix details the coverage summarized in
Section~\ref{sec:coverage}.

\paragraph{Composition.}
The 50 core tasks comprise ten per track, 33 adapted and 17 original.
They split 20/8/22 across 2D, 2.5D, and 3D and 12/13/8/8/9 across
puzzle, action, RPG, simulation, and narrative games.
Table~\ref{tab:stats} gives the mean structure of a task on each track,
and Table~\ref{tab:families} groups the tasks into five families by the
stressor each is designed to expose.

\paragraph{Difficulty annotation.}
Annotators score each task from 1 to 3 on four axes:
\emph{environment} (engine weight, spatial form, scene count, asset
authoring load), \emph{interaction} (input modalities, action families,
actions that must be coordinated), \emph{state} (branching, cross-scene
dependency, persistence, depth of required transitions), and
\emph{presentation} (camera, HUD, animation, feedback density). The tier
follows the mean over the four axes: easy up to 1.5, medium up to 2.3,
and hard above. Every track carries three easy, five medium, and two
hard tasks; the three tiers average 30, 48, and 71 leaf checks and 18,
36, and 76 journey steps.

\begin{table}[H]
\centering
\begin{minipage}{0.78\textwidth}

\caption{Mean task structure. Sc./Ent.: scenes/entities;
States: state nodes; Steps: reference-play inputs;
Leaves: leaf checks.}
\centering
\footnotesize
\setlength{\tabcolsep}{2.5pt}
\renewcommand{\arraystretch}{1.12}
\rowcolors{2}{LinkPanel}{white}

\begin{tabular*}{\linewidth}{
  @{\extracolsep{\fill}}lccccc@{}
}
\toprule
\textbf{Engine}
& \textbf{Sc.} & \textbf{Ent.}
& \textbf{States} & \textbf{Steps} & \textbf{Leaves} \\
\midrule
Defold        & 2.1 & 11.8 & 7.4 & 31.0 & 40.8 \\
Godot         & 2.3 & 13.1 & 8.1 & 35.0 & 45.1 \\
Roblox Studio & 2.6 & 14.9 & 8.6 & 41.0 & 48.0 \\
Unity         & 2.8 & 15.6 & 8.9 & 41.0 & 49.5 \\
Unreal Engine & 3.2 & 16.4 & 9.3 & 45.0 & 52.6 \\
\midrule
\rowcolor{black!6}
\textbf{All} & 2.6 & 14.4 & 8.5 & 38.6 & 47.2 \\
\bottomrule
\end{tabular*}

\label{tab:stats}
\end{minipage}
\end{table}

\begin{table}[H]
\caption{Task families and the stressor each family is designed to
expose.}
\centering
\small
\begin{tabularx}{\linewidth}{@{}P{0.20\linewidth}YY@{}}
\toprule
Family & Examples & Stressor \\
\midrule
T1 UI and rule puzzle & Four Shot, Arrow Escape          & authored content, event rules, input edges, reset \\
T2 Action             & Driftfolk Chronicle, Bomb Frog   & collision, timing, camera, feedback \\
T3 3D spatial action  & Dungeon Duel, Noir Roombreak     & navigation, combat state, multi-room transitions \\
T4 Systems            & Relay Foundry, Ape Tower Defense & resources, AI, dependency graphs, persistence \\
T5 Narrative          & There Are Signs                  & branch tables, authored presentation, visual assembly \\
\bottomrule
\end{tabularx}
\normalsize

\label{tab:families}
\end{table}
\section{Game Completion Boost: Skill and Evaluation Protocol}
\label{app:recovery}

\subsection{Artifact and invocation}
GCB contains a \texttt{SKILL.md} entry point, reusable
instructions, a read-only project-discovery script, and
standard-library Python helpers for checkpoints and repair trials. A Codex plugin manifest
registers the skill; other tool-capable agents can read the same
instructions. The agent's scaffold supplies engine access, input, capture, and image reading. The protocol starts a new local record from public requirements
for each task and keeps the skill fixed across tasks.

\subsection{Checkpoint graph and observation records}
The agent defines each checkpoint's public source, expected behavior,
input, prerequisites, relevant resources, and required evidence. The
helper validates unique identifiers and an acyclic dependency graph,
instantiates repeated visits as separate checkpoints, and schedules only
checks whose prerequisites currently pass, giving observed failures
priority over new checks on the same path.

An observation links a checkpoint to an artifact revision, process
session, input event, and outcome. A visual transition requires distinct
before/after captures that were actually shown to the model, and a
static visual check requires an after image; available public state or
logs are paired with these views, and an explicit comparison records
whether state and display agree. A check that lacks the required access
is marked unverified or tool-blocked. The helper verifies field coverage,
rejects captures reused across revisions, artifacts, or sessions, and
applies the same revision and session checks to final journey inputs and
reopen events.

\subsection{File monitoring, repair scope, and replay}
GCB binds its ledger to the assigned public project. On each status,
sync, or record call, it hashes regular files, including added and
deleted paths, with fixed cache exclusions documented in the package;
symlinks in monitored paths are rejected, and records stay outside the
project. File-to-resource patterns and shared resources are declared in
the contract. A local change invalidates the mapped checkpoints and their
transitive dependents, whereas a shared or unmapped change, including a
manual patch report that touches a shared resource, invalidates all
checks; every change also clears final replay.

A status call that finds unrecorded changes requests a sync, which saves
the invalidation, and a record call rescans before accepting an
observation and rejects it if new changes appear, so every observation
names the current fingerprint. The agent then relaunches or reloads the
build and repeats the affected inputs; a required reopen records distinct
process identities and the observed entry or saved state. Retries end at
the public budget, a persistent capability block, or two targeted repairs
that fail without new diagnostic evidence. The helper records each check
as observed pass, observed failure, unverified, or tool blocked; its
completion flag marks a complete record, and the hidden evaluator scores
the submitted game independently.

\paragraph{Rollback-aware self-refinement.}
Once a checkpoint passes, a repair trial saves the monitored files and
fixes the passed checkpoints as protected controls, and the agent names
an unresolved target and a repair hypothesis. A candidate is retained
only after fresh observations confirm that the target and every protected
check pass under the candidate revision, including checks that selective
replay would otherwise keep. If the target or a protected check is
observed to fail, GCB restores the saved state; missing or blocked
evidence leaves the trial inconclusive, and the agent may continue or
abandon it within the budget. All decisions use public observations only.

Before restoring, the helper archives the rejected candidate, advances
the revision, and invalidates all passes; the agent reloads the restored
build and repeats the required inputs. Later decisions return a short
diagnosis and an alternative strategy, following within-task reflective
feedback~\citep{shinn2023reflexion,madaan2023selfrefine}, and an
identical failed hypothesis for the same target is rejected.
Instructions and model weights stay fixed, and memory resets between
tasks. Snapshots share the monitor's file scope; a fingerprint check and
a persistent recovery marker detect changed candidates and interrupted
restoration, and pending recovery blocks finalization. Snapshotting,
replay, reflection, and restoration all consume the task budget.

\begin{table}[H]
\centering
\begin{minipage}{0.80\textwidth}

\caption{Execution stages of GCB.}
\centering
\footnotesize
\begin{tabularx}{\linewidth}{@{}P{0.19\linewidth}X@{}}
\toprule
Stage & Required action \\
\midrule
Specify & Map public requirements, prerequisites, and resources. \\
Build & Complete the required path; launch early. \\
Inspect & Link real input to state and image observations. \\
Repair & Archive a repair anchor; invalidate affected checks. \\
Replay & Recheck target and protected checks; retain or restore. \\
Finalize & Play the full journey; reopen when required. \\
\bottomrule
\end{tabularx}

\label{tab:actions}
\end{minipage}
\end{table}

\subsection{Evaluation scope and protocol}
The evaluation of Section~\ref{sec:res-method} uses Codex CLI with
GPT-5.6 Sol, GPT-5.6 Terra, and DeepSeek-V4-Pro on 15 high-difficulty
games, three per engine track, fixed before any GCB run
(Appendix~\ref{app:tasks}). Each comparison fixes the task, native
prompt, tool catalog, and 3600\,s budget. Of the three native runs of
each game, the baseline is the one with the highest $O$, so every gain
is measured against the strongest available run.

The evaluation follows the protocol released for full comparisons, which
fixes public inputs, model, reasoning setting, engine, tools, and budget
and gives both conditions the same image tools, so a gain reflects how
the agent uses its tools. The GCB condition loads the fixed skill and
its file-monitoring helper, whose bookkeeping, inspection, snapshots,
and restoration consume the normal budget. Each GCB run records tool
availability, resource use, checkpoint history, captures shown to the
model, detected file changes, invalidations, replays, and restorations.
\subsection{Relation to existing game agents}
GameDevBench studies image/video feedback for Godot
editing~\citep{chi2026gamedevbench}. OpenGame accumulates templates and
debug protocols~\citep{jiang2026opengame}; PlayCoder separates GUI
testing and repair~\citep{peng2026playcoder}; Recursive Game Creator
reviews candidate versions using gameplay and visual
traces~\citep{chen2026recursivegamecreator}.
Feedback-driven refinement and environment interaction also inform
our instructions~\citep{shinn2023reflexion,madaan2023selfrefine,wang2023voyager}.

GameGen-Verifier extracts specification-derived keypoints, uses state
and visual observations, and propagates failures through
dependencies~\citep{jia2026gamegenverifier}. Its runtime state injection
enables independent local checks. GCB instead schedules checks within
input-driven player paths during construction, followed by a final
journey. Like BeforeDone~\citep{luo2026beforedone}, GCB binds
verification records to file scopes and invalidates them after relevant
edits; Ekstazi selects regression tests from measured file
dependencies~\citep{gligoric2015ekstazi}, whereas GCB uses
agent-declared resource maps. GCB combines these established ideas in a reusable
intervention for the benchmark's game-building tasks, and
Section~\ref{sec:res-method} reports its evaluation.

\section{Experimental Configurations}
\label{app:configs}

This appendix lists the settings that Section~\ref{sec:setup} holds
fixed, so that every run can be reproduced.

\paragraph{Configurations.}
Table~\ref{tab:configs} gives the scaffold version, model identifier,
and reasoning effort of the seven configurations, all run at maximum
effort; every other scaffold option stays at its release default. All
configurations share each track's tool catalog (the command-line,
editor-automation, and capture tools of its tool manual) and the same
task prompt (Appendix~\ref{app:prompts}), and none receives
engine-specific hints or extra budget.

\begin{table}[H]
\caption{Scaffold versions, model identifiers, and reasoning-effort
settings. DS Harness is DeepSeek Harness.}
\centering
\footnotesize
\setlength{\tabcolsep}{2.2pt}
\renewcommand{\arraystretch}{1.14}

\begin{tabularx}{\columnwidth}{
  @{}P{0.22\columnwidth} c P{0.17\columnwidth} Y c@{}
}
\toprule
\textbf{Scaffold} & \textbf{Version}
& \textbf{Model} & \textbf{ID} & \textbf{Eff.} \\
\midrule

\rowcolor{black!3}
& & Astra
& \texttt{gpt-6-astra} & \texttt{max} \\

\rowcolor{black!3}
& & Sol
& \texttt{gpt-5.6-sol} & \texttt{max} \\

\rowcolor{black!3}
& & Terra
& \texttt{gpt-5.6-terra} & \texttt{max} \\

\rowcolor{black!3}
\multirow{-4}{*}{Codex CLI}
& \multirow{-4}{*}{\texttt{0.159.2}}
& V4-Pro
& \texttt{deepseek-\allowbreak v4-pro} & \texttt{max} \\

\midrule
DS Harness & \texttt{0.2.0-rc.2} & V4-Pro
& \texttt{deepseek-\allowbreak v4-pro} & \texttt{max} \\

\rowcolor{black!3}
Claude Code & \texttt{2.1.285} & Opus 5
& \texttt{claude-opus-5} & \texttt{max} \\

Kimi Code & \texttt{2.1.1} & Kimi K3
& \texttt{k3} & \texttt{max} \\

\bottomrule
\end{tabularx}

\label{tab:configs}
\end{table}

\paragraph{Runs and budget.}
Every run starts from a clean copy of its frozen starter in an isolated
project directory, on a host with the track's locked engine and editor
installed (Appendix~\ref{app:engines}); hosts run Windows 11 Home
(version 25H2, build 26200.9457) on machines equipped with an Intel
Core i7-14700KF CPU (20 cores, 28 threads), 64\,GB of RAM, and an
NVIDIA GeForce RTX 4070 GPU with 12\,GB of VRAM. In the main study, each game--configuration pair
is run three times, and scores are averaged over the three runs. The
3600\,s budget is measured in wall-clock time and includes compilation,
import, and play sessions, and token counts sum the input (including cached) and output tokens that each provider reports over all calls of a run. About 90\% of runs deliver before the limit, so for most runs
the budget does not bind. In the remaining runs, the agent is notified
ten minutes before the limit, the run is stopped at 3600\,s, and the
project is evaluated as it stands. These runs stall or loop and average only 16.3 points of $F$: the limit
stops unproductive runs rather than cutting nearly finished games short.

\section{Prompts}
\label{app:prompts}

This appendix gives the two prompts that shape every score: the task
prompt from which each game is built and the prompt of the D6 visual
judge. The task prompt template uses task-specific file names.

\paragraph{Task prompt.}
Every configuration receives the same task prompt on every track
(Listing~\ref{lst:task-prompt}). It asks for the complete game rather
than a prototype, points to the GDD, its state interface (the
observation and session services of the prompt), and the tool manual,
states the time budget, and places hidden material off limits. It names
no engine and no engine-specific technique; the engine enters only
through the starter and the tool manual. Swap runs
(Appendix~\ref{app:transfer}) receive a variant that names the target
engine, requires the frozen requirements to be kept, places the native
implementation and its source game off limits, and asks the agent to
report compile, build, and gameplay checks separately.

\begin{lstlisting}[style=prompt,
  caption={Task prompt for native runs, identical for every configuration and track.},
  label={lst:task-prompt}]
HARD WALL-CLOCK LIMIT: You have at most 3600 seconds for this implementation turn. At the deadline, the launcher will stop the process and retain the project exactly as it exists. There is no extension or second generation turn.

Build the complete playable game specified in GDD.md, together with all public observation and session services required by the task. Deliver the full requested product, not a prototype, an isolated mechanic, or a plan for future work.

Before editing, read README.md and every public document it indexes, including the complete game design document, toolchain instructions, asset catalogs, manifests, and applicable interface contracts.

Follow the numbered public requirements. Implement the game so that it can be played through the ordinary controls specified in the GDD. Your internal architecture, class names, and implementation strategy remain your choice.

Where session services are required, they must operate on the actual game: legal entry, capture, restoration, and object queries must perform the published behavior. Interface signatures, success flags, or fabricated state do not substitute for working functionality. Read-only observation interfaces must report the actual game state; they must not drive gameplay or manufacture outcomes.

Use the installed engine and tools specified in TOOLCHAIN.md. Use the supplied assets and preserve protected content. Keep changes within the declared writable submission area, and save the final scene or entry point at the required path.

Development tools and public smoke checks are optional aids that you may use within the time budget. Passing them does not certify the complete game or all required service behavior.

Work only within the supplied project and permitted resources. Do not inspect sibling projects, private reference implementations, hidden evaluator files, historical results, or hidden answer code.

Deliver the saved, runnable project before the deadline. Do not claim that a test passed unless you actually executed it.
\end{lstlisting}

\paragraph{Visual judge prompt.}
At each registered reference state, the D6 judge receives the reference
capture, the submission's capture, and the state label, and returns
five item scores, their total, and a short reason
(Listing~\ref{lst:judge-prompt}). It is told to score only what the two
images show, so gameplay correctness, code, and internal state never
enter D6; Appendix~\ref{app:stages} describes how the totals become
leaf credits.

\begin{lstlisting}[style=prompt,
  caption={Prompt of the D6 visual judge.},
  label={lst:judge-prompt}]
You are a strict reviewer of game visuals.

You will receive two images:
- Image 1: the reference game image (Reference).
- Image 2: the candidate game image (Candidate).
- Current state: <a short label, such as menu, initial gameplay, later gameplay, management interface, success, or failure>.

Evaluate the candidate's visual assembly and presentation relative to the reference using only content directly visible in the images.

Do not evaluate or infer:
- whether the gameplay is correct;
- whether buttons are actually interactive;
- the correctness of numerical values, economy, collision, damage, win/loss conditions, or save logic;
- code, scene nodes, logs, or internal state;
- anything not shown in the images.

The task provides the assets. Do not evaluate originality. Evaluate whether the assets have been assembled correctly, completely, and coherently into the final presentation.

Pixel-perfect matching is not required. Differences in moving-object positions, animation frames, or minor capture timing must not be penalized on their own. Reasonable equivalent layouts are allowed.

Penalize clearly missing elements, incorrect substitutions, misalignment, unintended occlusion, stretching, cropping, overlapping elements, placeholder assets, debug markings, and broken rendering.

Assign five scores, each from 0 to 2, for a total from 0 to 10. Each component must be one of: 0, 0.5, 1.0, 1.5, or 2.0.

1. Scene completeness and asset assembly: Are the main background, characters, objects, HUD, and applicable result interface complete? Are there missing elements, incorrect substitutions, or extraneous debug elements?
2. Composition and layout: Are the positions, spacing, alignment, visual balance, and interface layout reasonable?
3. Scale, layering, and clarity: Are element sizes, foreground/background ordering, cropping, and occlusion appropriate? Is important information readable? Do not judge whether that information is correct for gameplay.
4. Style and color coherence: Are asset styles, colors, lighting, silhouettes, and visual language coherent? Are there obvious visual conflicts?
5. Overall finish: Are there visible stretching, transparency errors, overlaps, blur, placeholders, or broken effects? Does the presentation resemble a deliverable game?

Use the following anchors consistently:
- 2.0: Comparable to or better than the reference quality, with no obvious problems.
- 1.5: Generally good, with only minor visible problems.
- 1.0: Basically complete, but with clear visual shortcomings.
- 0.5: Severely incomplete, disorganized, or visually poor.
- 0.0: Major content is missing, rendering is broken, or meaningful comparison is not possible.

Return only the following seven lines:

Scene completeness and asset assembly: <score>/2
Composition and layout: <score>/2
Scale, layering, and clarity: <score>/2
Style and color coherence: <score>/2
Overall finish: <score>/2
Total: <sum of the five scores>/10
Reason: <at most two sentences describing only directly visible differences between the images>
\end{lstlisting}

\section{Cross-Group Engine Swaps}
\label{app:transfer}

This appendix details the swap experiment behind
Table~\ref{tab:transfer-summary-mockup} and
Section~\ref{sec:res-swap}.

\paragraph{Protocol.}
A swap instance re-issues the frozen contract of a core task on an
engine of the other group. For each of the six heavyweight--script-centric
pairs, the five hardest games of each engine under the annotation of
Appendix~\ref{app:tasks} are re-issued on the other engine, so every
pair is swapped in both directions. Selecting the hardest games limits
ceiling effects on the heavyweight engines, where easier games would
leave little room to separate engines and configurations. The GDD body, state-interface keys, leaf families, family
budgets, and scorecard are unchanged; only the engine names, the
starter, and the tool manual follow the target track, while the scene
model, language, input API, asset pipeline, and build path change
completely. The target starter follows the same structure as a native
starter of its track, and native and swap runs are scored by the same
evaluator revision, each through the adapter of its engine. No source
code crosses engines, and the agent never sees the native solution. Swap
runs use the four Codex CLI configurations and DeepSeek Harness, with
the same budget and tool catalog as native runs.

\paragraph{Aggregation.}
For each game and configuration scored on both engines, the engine
effect is $\Delta F=F_H-F_S$, the functional score on the heavyweight
engine minus that on the script-centric engine, so positive values
favor the heavyweight engine whichever engine is native. Effects are
averaged over configurations within a game and then over games with
equal weight, and Table~\ref{tab:transfer-summary-mockup} pools all swap
instances of each engine pair. The subscripts in the table are standard deviations over the games of each engine pair. Because each pair swaps
five games in each direction, both of its engines receive the native
prompt and the swap variant equally often, and any advantage a game
draws from its native engine enters $\Delta F$ with opposite signs in
the two directions.

\paragraph{Case studies.}
Two swapped games show that the engine effect is not a fixed offset but
depends on the configuration. When \emph{Coolant Cargo} moves from
Roblox Studio to Unreal Engine, every configuration scores higher, in
line with the heavyweight advantage of Section~\ref{sec:res-swap}, yet
the gain ranges from 7.0 points for Codex CLI with GPT-6 Astra (83.0 to
90.0) to 77.9 for DeepSeek Harness (10.0 to 87.9), the largest shift in
the swap experiment. The uneven gains reorder the field: Codex CLI with
DeepSeek-V4-Pro falls from second to fourth, and DeepSeek Harness
climbs from fourth to third. On \emph{Spire Ascent}, moved from Godot to Unreal Engine, the two
scaffolds that share DeepSeek-V4-Pro trade places: Codex CLI leads by 17.3 points on Godot (47.3 against 30.0),
and DeepSeek Harness by 8.8 on Unreal Engine (87.3 against 78.5). With
the game, its contract, and the model held fixed, which scaffold wins
still depends on the engine. 

\section{Supplementary Results}
\label{app:results}

This appendix analyzes the runs further, first through the engine cells
of Table~\ref{tab:main} and then through construction traces and
delivered games.

\begin{table}[H]
\centering
\begin{minipage}{0.84\textwidth}

\caption{Summary derived from Table~\ref{tab:main}.
D1 and Play normalize integration and gameplay scores;
Gap is their difference. Low is the share of configurations
below 50\% gameplay, and Spread is the range of $F$.}
\centering
\footnotesize
\setlength{\tabcolsep}{2pt}
\renewcommand{\arraystretch}{1.12}
\rowcolors{2}{LinkPanel}{white}

\begin{tabular*}{\linewidth}{
  @{\extracolsep{\fill}}lcccccc@{}
}
\toprule
\textbf{Engine}
& \makecell{\textbf{D1}\\(\%)}
& \makecell{\textbf{Play}\\(\%)}
& \makecell{\textbf{Gap}\\(pp)}
& \makecell{\textbf{Mean}\\$\boldsymbol{F}$}
& \makecell{\textbf{Low}\\(\%)}
& \textbf{Spread} \\
\midrule
Defold        & 88.6 & 37.2 & 51.4 & 38.6 & 71.4 & 56.9 \\
Godot         & 88.7 & 23.4 & 65.3 & 27.6 & 85.7 & 35.3 \\
Roblox Studio & 91.3 & 35.7 & 55.6 & 37.7 & 85.7 & 53.6 \\
Unity         & 97.0 & 56.9 & 40.1 & 55.2 & 42.9 & 29.0 \\
Unreal Engine & 94.6 & 30.4 & 64.1 & 33.7 & 85.7 & 51.3 \\
\bottomrule
\end{tabular*}

\label{tab:engine-summary}
\end{minipage}
\end{table}

\paragraph{Engine-level summary.}
Table~\ref{tab:engine-summary} summarizes each engine over its
configurations. Integration stays between 89\% and 97\% of its points
on every engine, whereas the share of gameplay points ranges from 23\%
on Godot to 57\% on Unity, so the gap between the two shares runs from
40 points on Unity to 65 on Godot. On Godot, Roblox Studio, and Unreal
Engine, six of seven configurations earn under half of the gameplay
points, and on Defold five of seven; on Unity, where configurations
also lie closest together, four of seven clear that bar.

\Needspace{170pt}
\begin{wraptable}{r}{0.48\textwidth}
\captionsetup{font=footnotesize}
\caption{Kendall's $\tau$ between engine-specific configuration
rankings, computed using $F$ from Table~\ref{tab:main}.}
\centering
\begingroup
\footnotesize
\setlength{\tabcolsep}{3pt}
\renewcommand{\arraystretch}{1.16}
\rowcolors{2}{LinkPanel}{white}

\begin{tabular*}{\linewidth}{
  @{\extracolsep{\fill}}lcccc@{}
}
\toprule
& \textbf{Godot}
& \textbf{Roblox}
& \textbf{Unity}
& \textbf{Unreal} \\
\midrule
Defold        & 0.52 & 0.52 & 0.62 & 0.43 \\
Godot         &      & 0.62 & 0.33 & 0.33 \\
Roblox Studio &      &      & 0.52 & 0.14 \\
Unity         &      &      &      & 0.43 \\
\bottomrule
\end{tabular*}

\label{tab:rank-agreement}
\endgroup
\end{wraptable}

\begin{wraptext}
\paragraph{Rank agreement between engines.}
Table~\ref{tab:rank-agreement} gives Kendall's $\tau$ between the
configuration rankings of every pair of engines. Agreement ranges from
0.14 for Roblox Studio and Unreal Engine to 0.62 for Godot and Roblox
Studio and for Defold and Unity, and stays at or below 0.52 in eight of
the ten pairs. Agreement is higher within an engine group than across
groups (mean $\tau$ 0.52 against 0.40), yet engines of the same group
still differ, with $\tau=0.43$ for the two heavyweight engines and
0.52--0.62 among the script-centric ones. Four pairs of adjacent
configurations differ by less than the 1.5-point run noise
(Section~\ref{sec:setup}): Sol and Claude Code, and the two
DeepSeek-V4-Pro scaffolds, on Godot; the two DeepSeek-V4-Pro scaffolds
on Unity; and Terra and DeepSeek-V4-Pro under Codex CLI on Unreal
Engine. However these pairs are ordered, no $\tau$ exceeds 0.71, so no
two engines agree.
\end{wraptext}

\begin{table}[H]
\centering
\begin{minipage}[t]{0.49\textwidth}
\vspace{0pt}\captionsetup{font=footnotesize}

\caption{Gameplay bottlenecks across $F$ terciles of the 35
configuration--engine cells (12, 12, and 11 cells).
Weakest reports bottleneck frequency, and Mean share normalizes
dimension scores by their maxima. The all-zero gameplay cell
(Kimi Code on Defold) enters Mean share but not Weakest.}
\centering
\begingroup
\footnotesize
\setlength{\tabcolsep}{1.1pt}
\renewcommand{\arraystretch}{1.16}
\rowcolors{3}{LinkPanel}{white}

\begin{tabular*}{\linewidth}{
  @{\extracolsep{\fill}}lcccccc@{}
}
\toprule
& \multicolumn{3}{c}{\textbf{Weakest (\%)}}
& \multicolumn{3}{c}{\textbf{Mean share (\%)}} \\
\cmidrule(lr){2-4}\cmidrule(l){5-7}
\textbf{Dimension}
& Low & Mid & Top
& Low & Mid & Top \\
\midrule

D2 Completeness
& \textbf{45.5} & 8.3 & 0.0
& 11.7 & 35.4 & 67.6 \\

D3 Mechanics
& 36.4 & \textbf{50.0} & 27.3
& 12.8 & 33.0 & 61.4 \\

D4 Lifecycle
& 9.1 & 8.3 & \textbf{45.5}
& 17.9 & 40.1 & 55.6 \\

D5 Feedback
& 9.1 & 33.3 & 27.3
& 17.2 & 35.4 & 60.6 \\

\bottomrule
\end{tabular*}

\label{tab:weakest-dim}
\endgroup
\end{minipage}
\hfill
\begin{minipage}[t]{0.49\textwidth}
\vspace{0pt}\captionsetup{font=footnotesize}

\caption{Within-engine differences in $F$ between model tiers
and between DeepSeek Harness and Codex CLI.}
\centering
\begingroup
\footnotesize
\providecolor{ogblue}{RGB}{31,119,180}
\providecolor{ogorange}{RGB}{230,120,30}
\setlength{\tabcolsep}{1.1pt}
\renewcommand{\arraystretch}{1.16}

\begin{tabular*}{\linewidth}{
  @{\extracolsep{\fill}}lccccc@{}
}
\toprule
\textbf{Contrast}
& \textbf{Defold}
& \textbf{Godot}
& \textbf{Roblox}
& \textbf{Unity}
& \textbf{Unreal} \\
\midrule

\multicolumn{6}{@{}l}{\textit{Model contrasts under Codex CLI}} \\

Astra $-$ Sol
& \textcolor{ogblue}{$+10.2$}
& \textcolor{ogblue}{$+28.3$}
& \textcolor{ogblue}{$+33.4$}
& \textcolor{ogblue}{$+6.5$}
& \textcolor{ogblue}{$+41.4$} \\

\rowcolor{black!3}
Sol $-$ Terra
& \textcolor{ogblue}{$+19.3$}
& \textcolor{ogblue}{$+5.1$}
& \textcolor{ogblue}{$+20.2$}
& \textcolor{ogblue}{$+22.5$}
& \textcolor{ogblue}{$+8.8$} \\

V4-Pro $-$ Sol
& \textcolor{ogorange}{$-26.7$}
& \textcolor{ogblue}{$+12.7$}
& \textcolor{ogblue}{$+10.3$}
& \textcolor{ogorange}{$-16.1$}
& \textcolor{ogorange}{$-9.9$} \\

\midrule
\multicolumn{6}{@{}l}{\textit{Scaffold contrast with V4-Pro}} \\

\rowcolor{black!3}
Harness $-$ Codex
& \textcolor{ogblue}{$+15.8$}
& \textcolor{ogblue}{$+1.2$}
& \textcolor{ogorange}{$-21.9$}
& \textcolor{ogblue}{$+0.6$}
& \textcolor{ogblue}{$+19.8$} \\

\bottomrule
\end{tabular*}

\label{tab:contrasts}
\endgroup
\end{minipage}
\end{table}

\paragraph{Weakest dimension by score level.}
Table~\ref{tab:weakest-dim} gives the numbers behind the progression of
Section~\ref{sec:res-gap}. The configuration--engine cells are split
into thirds by $F$, and each gameplay dimension is expressed as a share
of its weight. In the lowest third, completeness is the weakest
dimension in nearly half of the cells, and completeness and mechanics
each earn about an eighth of their weight. In the middle third,
mechanics is weakest in half of the cells, ahead of feedback. In the top
third, completeness, mechanics, and feedback each pass 60\% while
lifecycle stays at 56\%, and lifecycle is most often the weakest. As $F$
rises, the weakest dimension thus moves from content to rules and then
to lifecycle.

\paragraph{Model and scaffold contrasts.}
Table~\ref{tab:contrasts} gives the contrasts of
Section~\ref{sec:res-rank} on every engine. Both model-tier contrasts
within the GPT family keep their sign on every engine, whereas the
cross-family contrast and the scaffold contrast each change sign.

\begin{table}[H]
\centering
\begin{minipage}{0.88\textwidth}

\caption{Action shares in the construction traces of native runs (\%).
Other covers reading files and documents, planning, and remaining
actions.}
\centering
\begingroup
\footnotesize
\color{black}
\setlength{\tabcolsep}{4pt}
\renewcommand{\arraystretch}{1.16}

\begin{tabularx}{\linewidth}{
  @{}ll*{5}{>{\centering\arraybackslash}X}@{}
}
\toprule
\multicolumn{2}{c}{\textbf{Configuration}}
& \multicolumn{5}{c}{\textbf{Workflow steps (\%)}} \\
\cmidrule(lr){1-2}\cmidrule(l){3-7}

\textbf{Scaffold}
& \textbf{Model}
& \textbf{Edit}
& \textbf{Build}
& \textbf{Play}
& \textbf{View}
& \textbf{Other} \\
\midrule

\cellcolor{black!5}
& GPT-6 Astra
& 26.3 & 17.4 & 7.7 & 5.6 & 43.0 \\

\rowcolor{black!3}
\cellcolor{black!5}
& GPT-5.6 Sol
& 17.1 & 19.9 & 5.3 & 4.8 & 52.9 \\

\cellcolor{black!5}
& GPT-5.6 Terra
& 31.3 & 22.4 & 5.0 & 4.2 & 37.1 \\

\rowcolor{black!3}
\cellcolor{black!5}\multirow{-4}{*}{Codex CLI}
& DeepSeek-V4-Pro
& 28.3 & 8.9 & 8.7 & 4.6 & 49.5 \\

DeepSeek Harness
& DeepSeek-V4-Pro
& 34.1 & 6.3 & 3.1 & 4.5 & 52.0 \\

\rowcolor{black!3}
Claude Code
& \claudemodel
& 20.3 & 7.5 & 0.2 & 2.2 & 69.8 \\

Kimi Code
& \kimimodel
& 20.3 & 9.6 & 0.0 & 0.0 & 70.1 \\

\bottomrule
\end{tabularx}

\label{tab:trace-composition}
\endgroup
\end{minipage}
\end{table}

\paragraph{What agents do with their steps.}
Table~\ref{tab:trace-composition} breaks the workflow steps of the
native runs down by kind: editing files and assets, building (import,
compile, and test), playing (launching the game or sending it input),
and viewing (a tool returns a game image to the model, or the agent
opens a captured frame with an image-reading tool). Saving an image
without exposing it to the model counts under Other. Playing and viewing
together take at most 13.3\% of any configuration's steps, 2.4\% for
Claude Code, and none for Kimi Code.

\Needspace{170pt}
\begin{figure}[H]
\centering
\setlength{\fboxsep}{0pt}
\shot{0.32\linewidth}{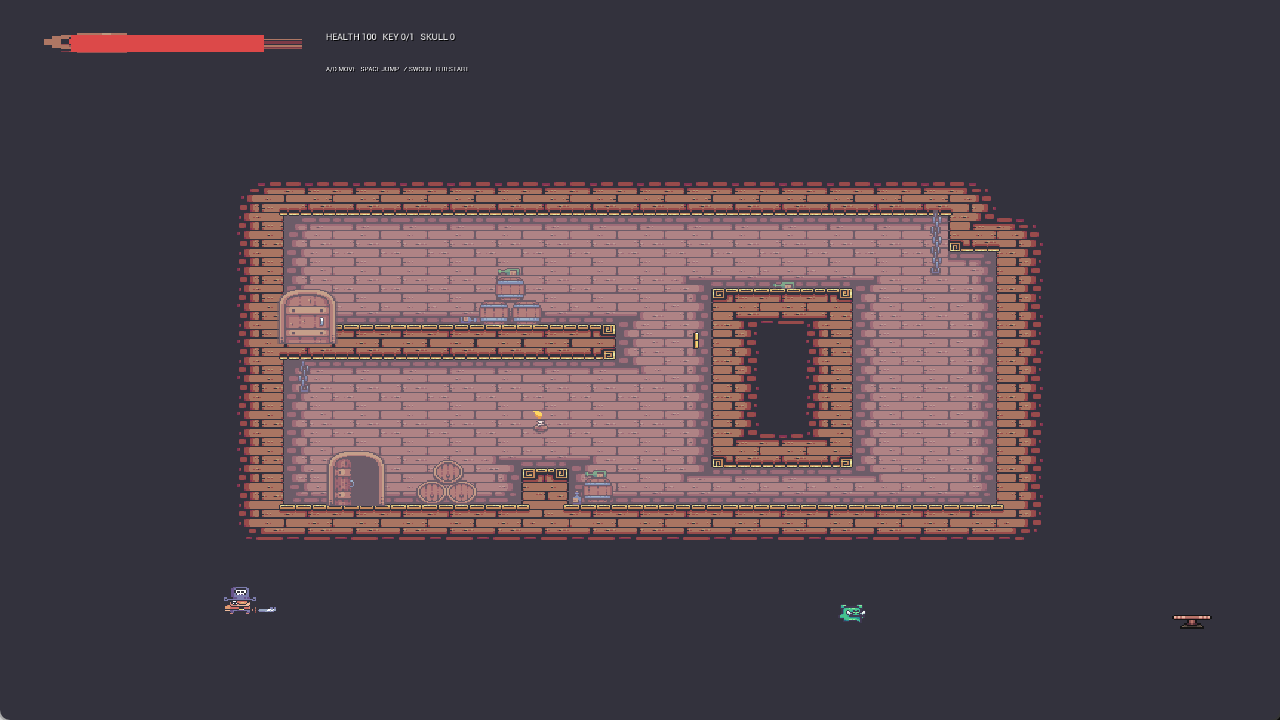}\hfill
\shot{0.32\linewidth}{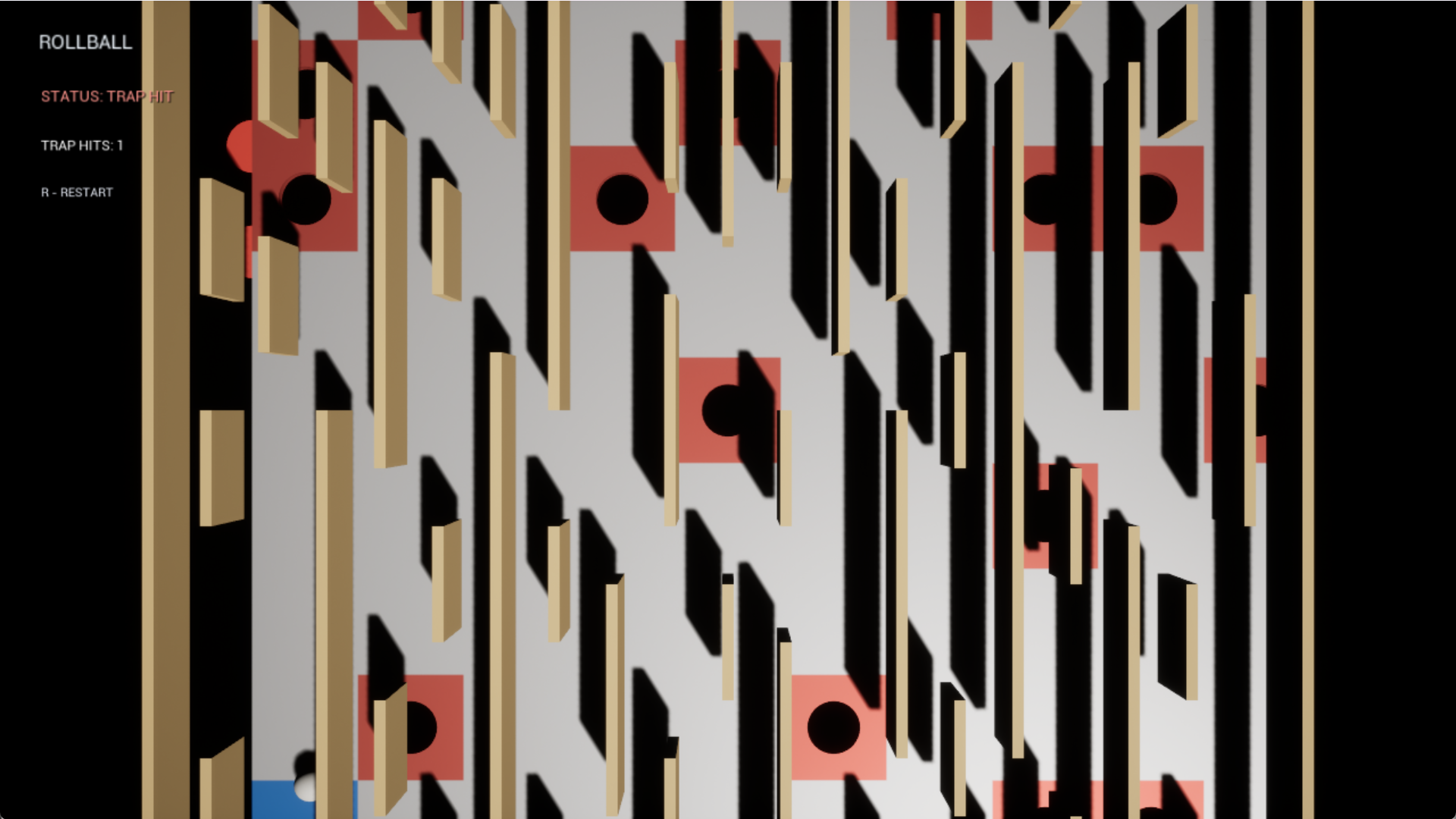}\hfill
\shot{0.32\linewidth}{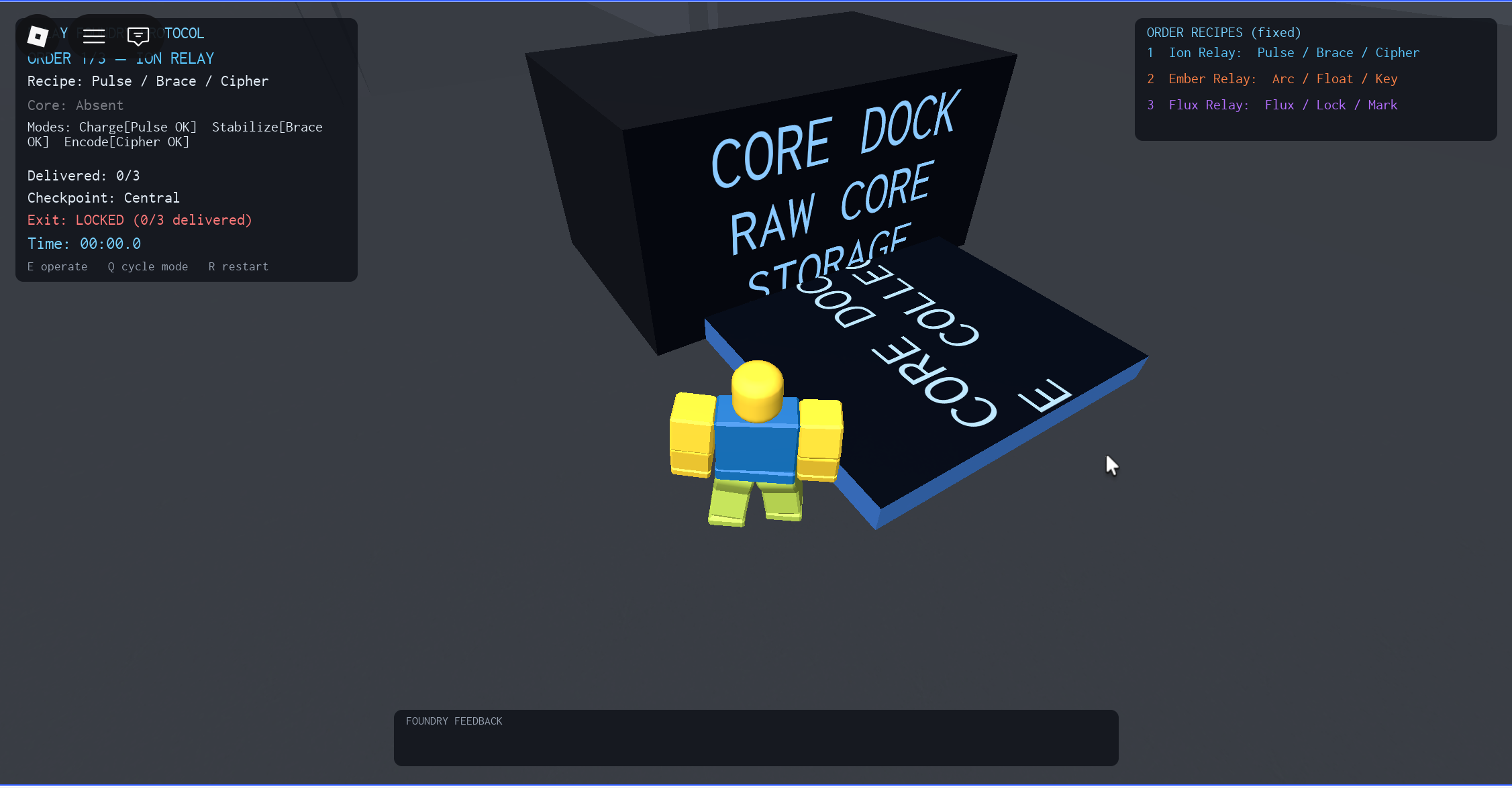}
\caption{Three delivered games from the bottom, middle, and top of the
score range, as captured by the evaluator. (a) \emph{Captain2D
Platformer} on Unreal Engine ($O=15.2$): the player character, enemies,
and springs stand below the room. (b) \emph{RollBall} on Unreal Engine
($O=57.2$): the HUD reports a trap hit while the cropped view leaves the
ball at the lower edge. (c) \emph{Relay Foundry} on Roblox Studio
($O=95.7$): the rules hold, but system buttons and the HUD crowd the
world signs.}
\label{fig:failure-cases}
\end{figure}

\paragraph{Failure cases.}
Three delivered games, from the bottom, middle, and top of the score
range, show in single runs how the weakest part of a game shifts with
capability (Section~\ref{sec:res-gap}; Figure~\ref{fig:failure-cases}).
In \emph{Captain2D Platformer} on Unreal Engine ($O=15.2$), the project
builds, $\iface$ publishes fresh, well-formed state, and the opening
health, key, and death counts are correct, yet the player character,
the enemies, and the springs are placed below the room, so the goal
cannot be reached with public controls. The objective leaves
behind it are therefore blocked by the product rather than observed to
fail, and the evaluator draws no conclusion about the key logic they
would have tested. In \emph{RollBall} on Unreal Engine ($O=57.2$), each
trap counts once and restart opens a fresh round within two seconds, but
the rules contradict the screen: holes are drawn away from the trap
triggers, a trap fires while the ball rests on visibly safe floor, and
the camera crops the starting ball and the exit. In \emph{Relay Foundry}
on Roblox Studio ($O=95.7$), all three orders complete through real
input, wrong phases and modes fault and purge as specified, and the HUD
tracks $\iface$; the points it loses lie in restart and in screen
layout, where system buttons and the HUD overlap the world signs. The
three games fail at content, rules, and lifecycle in turn, the same
progression that Table~\ref{tab:weakest-dim} shows across terciles.

\end{document}